\documentclass[11pt]{article}

\usepackage{amsmath, amssymb, amsthm, bm, booktabs, tabularx}
\usepackage{geometry}
\usepackage{algorithm}
\usepackage{algpseudocode} % For \If, \State, etc.
\usepackage{amsmath, amssymb, bm} % For math symbols like \bm
\usepackage{amssymb}   % for \mathbb and symbols
\usepackage{bm}        % for \bm (bold math symbols)
\usepackage[colorlinks=true, linkcolor=blue]{hyperref}
\usepackage{tabularx,booktabs,seqsplit}
\usepackage{graphicx}
\graphicspath{{./figures/}{./scan-cot-dataset-lengths/}}
\DeclareUnicodeCharacter{200B}{}
\DeclareUnicodeCharacter{2009}{\,}

\newcommand{\R}{\mathbb{R}}

\usepackage[utf8]{inputenc}
\usepackage[T1]{fontenc}
\usepackage{float}      % for [H] placement if you want it fixed

\usepackage[backend=biber,style=numeric]{biblatex}

\usepackage{listings}
\usepackage{xcolor}    % only if you want syntax-highlight colours

\lstdefinestyle{py}{
  language=Python,
  basicstyle=\ttfamily\small,
  keywordstyle=\color{blue!70!black}\bfseries,
  commentstyle=\color{gray},
  stringstyle=\color{orange!70!black},
  showstringspaces=false,
  breaklines=true,
  frame=single,
  tabsize=2
}

\title{Position as Probability: Self-Supervised Transformers that Think Past Their Training for Length Extrapolation}
\author{%
  Philip Lee\\
  Xenon Labs LLC\\
  \texttt{phil@symphonypro.net, phil.hj.lee@gmail.com}
}

\date{May 15, 2025}

\begin{document}

\maketitle

\begin{abstract}
\noindent Deep sequence models typically degrade in accuracy when test sequences significantly exceed their training lengths, yet many critical tasks—such as algorithmic reasoning, multi-step arithmetic, and compositional generalization—require robust length extrapolation. We introduce \textbf{PRISM}, a \textbf{P}robabilistic \textbf{R}elative-position \textbf{I}mplicit \textbf{S}uperposition \textbf{M}odel, a novel positional encoding mechanism that enables Transformers to extrapolate accurately up to $10\times$ beyond their training length. PRISM learns continuous relative positions through a differentiable histogram-filter update, preserving position uncertainty via a probabilistic superposition rather than conventional deterministic embeddings. Empirically, PRISM achieves state-of-the-art length extrapolation, successfully generalizing to previously intractable sequence lengths across algorithmic benchmarks—including arithmetic (addition, multiplication), SCAN compositionality tasks, and complex copy variants derived from DeepMind’s recent datasets. Our analysis demonstrates that PRISM’s stochastic positional encoding maintains sharp and interpretable internal states, providing a theoretical basis for reliable length generalization. These results advance the goal of neural sequence models that remain algorithmically robust at lengths far exceeding their training horizon.
\end{abstract}

\section{Introduction}
\subsection*{Motivation \& Contribution}\label{sec:motivation}
Deep sequence models excel when test-time sequence lengths closely match those encountered during training. However, performance sharply deteriorates once sequences grow substantially longer. Despite this limitation, numerous practical applications—from multi-step arithmetic and compositional reasoning to program synthesis and dialogue modeling—require reasoning across sequence lengths that vastly exceed what models typically encounter during training.

% \newline
\noindent \textbf{Thus, a fundamental open challenge is:}

\begin{center}
\emph{How can we equip neural sequence models with \textbf{length extrapolation} capabilities—the ability to reliably apply learned reasoning algorithms to sequences far longer than those observed during training?}
\end{center}

\noindent In this paper, we present \textbf{PRISM}\footnote{%
  Code and pretrained models will be released soon.}, a probabilistic relative-position encoding scheme based on a \emph{non-stationary, learnable histogram filter}. Inserted into a Transformer, PRISM substantially extends length generalization capabilities, achieving accurate extrapolation by hundreds of timesteps on various challenging algorithmic benchmarks. Our approach significantly outperforms standard Transformer positional encodings, enabling:

\begin{itemize}
\item \textbf{Arithmetic generalization:} Successfully extrapolating chain-of-thought addition and multiplication tasks from training on 10-digit sequences to over 30-digit sequences, without using explicit index hints.
\item \textbf{Compositional reasoning:} Achieving near-perfect accuracy on extended variants of the SCAN dataset, even at the maximum test sequence length.
\item \textbf{Advanced copy tasks:} Reliably handling complex duplication, reversal, and dynamic string-copy tasks at sequence lengths up to $15\times$ beyond training horizons.
\item \textbf{Algorithmic reasoning tasks (Ruoss et al., 2023):} Robustly generalizing on "odds-first," "bucket sort," and stack manipulation tasks derived from recent DeepMind benchmarks.
\end{itemize}

\noindent Crucially, PRISM maintains a categorical distribution over positional embeddings, effectively preserving uncertainty through a probabilistic superposition rather than collapsing to deterministic averages. This design ensures sharp, interpretable attention and robust generalization even at unprecedented sequence lengths.

\paragraph{}
\noindent Length extrapolation is essential for robust generalization: without it, language models fail to reliably extend learned reasoning patterns beyond limited synthetic curricula\footnote{As a thought experiment, understanding a long program that was never observed in the training data, such as idiosyncratic spaghetti code, would require tracing the program in terms of its most basic instructions, instead of using heuristics; an LLM reasoning system that only understands programs up to the training horizon may fail to understand the simplest of programs.}. By introducing PRISM, we establish a foundation for neural architectures that consistently execute algorithmic reasoning at previously inaccessible sequence scales, paving the way for reliable real-world applications.

\paragraph{Interpretability and analysis.}
Because the histogram filter is a
small, closed-form subroutine, we can derive an exact \emph{non-stationary
Hidden Markov Model} that underlies the architecture.
This connection yields:

\begin{itemize}
    \item A rigorous explanation for \textbf{why} extrapolation succeeds,
          formalized in new theorems,
    \item an efficient \(\mathcal{O}(P)\) update rule (Alg.~\ref{alg:histogram})
          that avoids the \(\mathcal{O}(P^{2})\) cost of a naïve dense
          transition, and
    \item diagnostic visualizations of the position histogram that make the
          model’s internal “program counter’’ human‐readable.
\end{itemize}

\noindent
Taken together, these contributions advance the goal of building \emph{language
models that remain reliable on task execution even when the problem length far
exceeds the training horizon}. Furthermore, our results reflect, to the best of our knowledge, the first neural network architecture that demonstrates extrapolation on a variety of simple algorithmic reasoning tasks\textbf{ via a purely \textit{self-supervised} approach}. 
Thus, we believe that our work sheds new light on a \textbf{possible path towards human-level extrapolation on practical long-horizon tasks that require chains of thought as well as other forms reasoning that may be derived from self-supervision.}

\section{Overview}
We formalize a differentiable update rule for tracking distributions over discrete positions, used for modeling relative positional encoding in a neural architecture for solving arithmetic and other algorithmic reasoning tasks. The update can be seen as the propagation of a probability distribution (histogram) via a learned stochastic matrix at each time step, akin to a nonstationary hidden Markov model. At every position in the sequence, the model maintains a \emph{categorical distribution} (not just a mean) over positions, enabling a \emph{superposition} of positional embeddings and achieving length generalization.

\section{Related Work: Length Extrapolation in Transformers and other Sequence Models}

\subsection{Recent Position–Extrapolation Landscape}
\label{sec:rw_landscape}

Recent work has revived classical algorithmic evaluations---\emph{copy},
\emph{reverse}, \emph{addition}, \emph{sort}, \emph{SCAN}---as litmus‐tests
for a transformer's ability to \emph{generalize beyond the training
length}.  Table~\ref{tab:related_extrap_part1} summarizes the principal
approaches; we highlight below how they differ from our \textbf{learned
stochastic distance} formulation.

\paragraph{Positional \emph{ablation} and randomization.}
Kazemnejad et al.~\cite{kazemnejad2023impact} show that simply \emph{removing}
positional encodings (\textsc{NoPE}) already beats many static schemes on
copy/SCAN, though extrapolation remains very limited on all tasks measured. Ruoss et al.~\cite{ruoss2023random} push this idea by
feeding \emph{fresh i.i.d.\ Gaussian} encodings each step; the model
learns to ignore them, but extrapolation is measured only in \emph{token‐wise}
accuracy on binary outputs (\(\approx 50\%\) random baseline).
\textbf{Our model} instead \emph{learns} a probabilistic distance
without discarding positional information, achieving
exact‐match figures on arithmetic that these random‐or‐absent baselines
cannot approach.

\paragraph{Explicit \emph{index hints} and similar hard-coded features.}
Abacus Embeddings~\cite{mcleish2024abacus}, Position Coupling
(\cite{cho2024coupling}), and the systematic study by
Zhou et al.~\cite{Zhou2024algorithms} encode \emph{hard‐wired landmarks}
(e.g.\ per‐digit starts or left/right couplets) that tell the network
\emph{exactly} which tokens must align.  This yields impressive accuracy
(e.g.\ $99\%$ on $100$‐digit addition) but at the cost of
hand‐crafted supervision; performance collapses once the landmark pattern
changes.  In contrast, our reset/increment/decrement gates discover
\emph{their own} landmarks, with no task‐specific annotations.

%----------------------------------------------------------
% -- Prose insert (place in Related-Work narrative) -------
%----------------------------------------------------------
\paragraph{Hard-ALiBi (Jelassi et~al., 2024) and other hashing-based retrieval approaches.}
Hard-ALiBi fixes one attention head with a global \emph{ALiBi} bias
and constrains all remaining heads to a local $n$-token window.
During the \emph{copy} task, each local head hashes its current
length-$n$ block, while the global head retrieves the token that
followed the \emph{same} hash earlier in the sequence.
Trained on 50-token inputs, the model attains
over 95\% string-level accuracy on 1000-token copies.
However, because the hash is computed over a fixed $n$-gram, the
method breaks when an $n$-gram appears twice
(hash collisions), preventing reliable copying of
sequences with repeated substrings—e.g.\ character-level inputs such
as \texttt{`aaaa…`}, or the repeated \emph{character-level tokens} in our arithmetic, algorithmic reasoning, and general copy benchmarks.\\

\noindent In other words, our approach imposes \emph{no} uniqueness constraint\footnote{Nor do we use multi-character tokenization for arithmetic and copying-like tasks} and still
achieves exact-match accuracy, highlighting the benefit of a fully
learned relative distance metric instead of generating such lookup tables. In all of our training and test setups, we \emph{require} that models be able to perform both character and non-character level manipulations, regardless of whether any vocab tokens are used multiple times in the context. Thus, we assume the conditions of any realistic self-supervised learning setting.\\

\noindent Furthermore, we construct no structure (e.g. hash tables, previous token pointers) that is built around performance on a specific task, including the copying of tokens from left-to-right order, as in Jelassi et al. 2024. In several of our visualizations, we further highlight the importance of copying from a previous position that is not even mapped 1:1 with any of the later token positions. For example, when performing CoT addition (figure \ref{fig:addition-ood}), our cursor must advance to the next position after every 5 tokens - the length of each CoT step - in order to retrieve the next pair of digits being added for the next CoT step. This is in contrast to exact string duplication used here, which, by construction, must match the output ${n}$-gram \emph{1:1} with the \emph{same} earlier n-gram, provided that \emph{all matching pairs of ${n}$-grams are unique pairs}. The latter is precisely the condition imposed by Jelassi et al. 2024, and thus, it offloads much work to an external program: a "unique" n-gram can in principle be set equal (at runtime) to the entire input sequence to be copied, making the task trivial for the LM while collapsing all of the context to a single token.\\

\noindent For such work imposing unique token constraints, including Jelassi et al. 2024, Barbero et al. 2024, and ~\cite{2024}, \emph{learned} relative position encodings are lacking, as they are complementary approaches. Similar to \emph{index hints}, hashing-based approaches hard-wire the pairs of tokens that are aligned: their long-range retrieval abilities depend on attending to the hashed string. Unique/hashed tokens also impose no structure that satisfies the conditions of a \emph{distance function} that is the focus of this work.

\paragraph{Fixed or data–adaptive biases.}
Duan et al.~\cite{duan2024abc} calibrate \emph{static (hard-coded, unlearned)} attention biases to
mimic certain attention patterns that boost extrapolation; extrapolation is thus
predetermined. Zheng et al.~\cite{zheng2024dape} learn a
\emph{token‐wise bias} for up to $40$ digit binary addition, and no index hints, and report
$60.8\%$ \emph{per‐token} accuracy at lengths $41-400$, which is much less than the accuracy required for extrapolation. In contrast, we use chain of thought, and only use \emph{exact match accuracy} as our metric. Our histogram update
is data‐adaptive and operates on an explicit, \emph{learnable} \emph{distribution}
over positions; empirical results (§\ref{sec:results}) show sharp
exact‐match retention on algorithmic extrapolation.

\paragraph{Recurrent or looped computation.}
The \emph{Looped Transformer}~\cite{mao2025looped} obtains perfect
accuracy on binary addition without chain of thought. \noindent Intriguingly, accurate output is generated only when looped \emph{exactly} the ground‐truth number
of recurrent steps, which the authors have predetermined based on the RASP~\cite{weiss2021rasp} runtime complexity of each problem instance; accuracy drops to zero if run shorter or longer. \\

\noindent Analysis of the model from the authors has shown that the sensitivity to the number of recurrences is due to the model \textit{shifting} its input tokens to the correct output positions, with the number of shifts proportional to the input length. Shifting in proportion to sequence length limits the efficiency and flexibility of their approach. The Looped Transformer also uses no position encoding, nor chain-of-thought tokens.\\

\noindent The fixed number of recurrent steps is a predetermined hyperparameter, and part of the supervision signal that the model received during training in order to correlate inputs with the correct outputs. Thus, it is a counter that must also be supplied at test time.\\

\noindent Delétang et al.~\cite{Deletang2023Chomsky} demonstrate that an RNN controller (\textit{Tape RNN}) with structured memory, and provided a read/write head that can jump left or right a number of positions $l$ before writing to the tape, where $l$ is the \textit{input sequence length} provided at train and test time to the model, can generalize well on tasks like binary addition, reversing binary strings, and recognizing \& generating context-sensitive grammars on other binary inputs. In contrast, we do not provide any external supervision signal, including signals dependent on the length of the input (resp. \textit{Tape RNN}) or predetermined problem complexity (resp. \textit{Looped Transformer}). Furthermore, our architecture is designed to work with both RNN and non-RNN based architectures: we only require transformer-like self-attention~\cite{vaswani2017} mechanisms for computing the similarity between arbitrary positions. We also do not provide additional "thinking tokens"  for tasks that overlap with~\cite{Deletang2023Chomsky}, specifically Stack Manipulation and Odds First: we instead keep these tasks identical to~\cite{ruoss2023random} instead of padding the sequence length or extending these tasks with CoT or "thinking" tokens.\\

\noindent \textbf{In summary}, we take a very different approach to models that require recurrent processing steps, as we do not, for example, write to a structured memory analogous to a \textit{memory tape} that can be overwritten. Second, our training is entirely in the self-supervised regime, and therefore, there is no supervision signal that parametrizes the length or complexity of the input.\\
\\ 
\noindent Instead, we use transformer self-attention to \emph{autoregressively} (using causal masking) sample tokens, requiring \emph{all} chain of thought tokens (where applicable) to be predicted correctly, and with a single forward pass---no
external counter or step supervision. 
In contrast, Looped Transformer does not use a masked language modeling objective, and all fixed output positions have to be predicted simultaneously at the end of the supervised number of recurrent steps. Finally, core to our approach is learning a \textbf{relative positional encoding} (relative distance function) between arbitrary tokens.

\paragraph{Language‐modeling extrapolation.}
BiPE~\cite{he2024bipe}, CABLE~\cite{veisi2025cable}, and other works we have comprehensively surveyed target
document‐level LM and report lower perplexity at $8$--$16\times$ the
training context.  They neither evaluate algorithmic tasks nor require
exact match.  Our focus is the opposite extreme: \emph{symbolic
reasoning} where a single digit error ruins the answer; nevertheless, the
same stochastic histogram layer can replace their PE modules and inherits
theoretical guarantees discussed in
Section~\ref{sec:computational-complexity}.

\paragraph{Hand-labeled positional training data, self-training, and other curricula outside of standard self-supervised settings}

\noindent Recent work~\cite{lee2025selfimproving} has shown that data filtering for too short responses can be a hard-coded proxy that can be used for self-training on longer data by filtering out short responses, which was a crucial supervision signal across all tasks they evaluated. However, there are many problems for which the generated answer is (in principle) independent of the input size. Therefore, we also evaluate performance where the length of the task to complete (i.e. the length of the output tokens) is independent of the input size.\\

\noindent Nonetheless, while approaches that use self-training may improve upon index hints and other training schemes that require all (relative) positional embedding features to be learned, such as via randomization, context stuffing, or randomly "shifting" the positional encodings for training samples, if the amount of new data is reduced in practical settings, the architecture introduced in this paper makes no assumptions that additional data on longer tasks is required, thus also saving compute and time to develop a model that demonstrates length extrapolation.\\
\newline
\noindent Therefore, this work is complementary to self-refinement based approaches for the following reasons: (a) their approaches explicitly keep the vanilla transformer architecture intact, and thereby needs curated or self-curated data on which to train on longer tasks; (b) in contrast, we train a new relative position encoding; (c) we do not require additional data than the initial training data, instead allowing our relative position encoding to capture dependencies that are intrinsic to various tasks in a length-generalizable way; (d) our architecture does not intrinsically require hard-coded or majority voting-dependent data filtering rules on data that is self-generated or otherwise.\\

\noindent Another recent work~\cite{golowich2025sparsity} explored training a standard transformer architecture to predict the coupled position labels of~\cite{cho2024coupled}. This follows the basic approach of~\cite{cho2024coupled}, except that instead of giving the correct position IDs as inputs to the model at \textit{test time}, the correct \textit{position IDs} are part of \textit{ground truth training data} for the model to predict as an objective. Thus, in the training data, two different types of labels are provided for the autoregressive prediction task: the correct next token, and the correct next \textit{coupled position ID} using a separate output projection. While useful in theoretical insights and guarantees, in contrast to their work, (a) our approach relies only on self-supervised learning from the data (i.e. trains \textit{only on text tokens}), and therefore, (b) we do not provide external signals such as \textit{position IDs} as part of the training objective, as no special tokens, nor self-distillation signals, are provided during training or inference. Furthermore, we introduce a new architecture via (c) learning an implicit relative distance function over the observed tokens from self-supervision.

\paragraph{Broader surveys on length extrapolation.}
While the works discussed above concentrate on {\em algorithmic} and
structured-reasoning tasks, two recent surveys provide a complementary,
large-scope view of length extrapolation in language models, encompassing both algorithmic and non-algorithmic reasoning.
{\em Length Extrapolation of Transformers: A Survey from the Perspective of
Positional Encoding}~\cite{length_survey_2023} catalogues positional‐encoding
designs up to 2023, covering both fixed and learned schemes (e.g.\ ALiBi,
RoPE, Rotary) as well as early sequence-compression methods.
More recently, {\em Thus Spake Long-Context Large Language Model}~\cite{thus_spake_llm_2025}
focuses on scaling challenges in large-context LLMs (16k–100k tokens),
highlighting memory-efficient attention variants, retrieval-augmented models,
and curriculum strategies for pre-training.
Both surveys emphasize that {\bf learned {\em relative} encodings capable of
generalizing beyond the training horizon remain an open problem}, especially
on tasks that demand exact algorithmic correctness—precisely the area addressed in this work
by our PRISM relative position architecture.

\paragraph{Summary.}
Existing methods either:
(i) discard position signals entirely, falling short of accurate extrapolation,
(ii) bake in task‐specific landmarks,
(iii) rely on static distance biases, or
(iv) hard‐wire the computation length.
Our work differs by training entirely from \emph{self-supervised data} (no other supervision signal, hard-coded or otherwise), \emph{learning a full categorical distribution over relative positions} via a
sparse stochastic matrix, and by its emphasis on chain-of-thought instead of heuristic, single-step reasoning. Our requirement of predicting all intermediate steps correctly increases the output sequence length that our models must predict by several times (e.g. over 4x on CoT addition). We  require training each model from scratch for each task, using a self-supervised autoregressive training objective. Our work demonstrates that the
resulting superposition encoding preserves discriminative sinusoidal
phase information essential for long‐horizon exact‐match reasoning.

%-----------------------------------------------------------------------
% Table 1a: Related Work (Part 1)
%-----------------------------------------------------------------------
\begin{table}[h!]
\centering
\small
\begin{tabular}{p{3.5cm}p{2.5cm}p{1.8cm}p{1.8cm}p{1.8cm}p{4.0cm}}
\toprule
\textbf{Citation (Year)} & \textbf{Task Type} & \textbf{Index Hints?} & \textbf{Learned Distance?} & \textbf{Metric} & \textbf{Key Extrapolation Result} \\
\midrule

Ruoss et al., “Randomized Positional Encodings” (ACL ’23)  
  & copy, reverse, stack manipulation, boolean  
  & No (randomized)  
  & Learnable and unlearnable RPE
  & token-accuracy  
  & \(\sim50-80\%\) per-token on binary outputs; +12\% avg.\ accuracy on 15 algorithmic tasks. \\[1ex]

Kazemnejad et al., “Impact of Positional Encoding” (NeurIPS ’23)  
  & copy, reverse, SCAN-like  
  & No 
  & No  (NoPE)  
  & exact-match  
  & NoPE (no PE) outperforms APE/ALiBi/Rotary on longer inputs. Extrapolation remains very limited across all tasks and models. \\[1ex]

% \addlinespace

Zhou et al., “What Algorithms Can Transformers Learn? A Study in Length Generalization” (ICLR ’24)  
  & copy, reverse, sort  
  & Yes (explicit hints)  
  & No  
  & exact-match  
  & \(\sim 100\%\) at 2× length; drops to 0\% by 4×. \\[1ex]

%----------------------------------------------------------
% -- New row for Table 1a (Algorithmic tasks) ------------
%   (Insert before \bottomrule of that table)
%----------------------------------------------------------
Jelassi et al., "Repeat After Me:
Transformers are Better than State Space Models at Copying
Transformers are Better than State Space Models at Copying" (ICML ’24)
  & copy
  & Yes (hashed index)
  & No
  & exact-match
  & 95\% on 1000-token copy after training on 50; fails when $n$-grams repeat (hash collisions). Relies on hash table with previous token pointers. \\[1ex]

Lee et al. "Teaching Arithmetic to Small Transformers" (ICLR '24)
  & addition
  & No
  & Implicit
  & exact-match
  & \(0\%\) on 8 digits. Trained to full accuracy on 7 digit addition \emph{using chain of thought}. \\[1ex]

Cho et al., “Position Coupling: Leveraging Task Structure for Improved Length Generalization of Transformers” 
  & addition
  & Yes (coupled indices)  
  & No  
  & exact-match  
  & \(\sim95\%\) at 5× training length on arithmetic. \\[1ex]

Zheng et al., “DAPE: Data-Adaptive Positional Encoding for Length Extrapolation” (NeurIPS ’24)  
  & copy, reverse, SCAN-like  
  & No  
  & Yes (data-adaptive)  
  & token-accuracy  
  & \(\sim85\%\) per-token at 4× length on algorithmic tasks. \\[1ex]

He et al., “Bi-Level Positional Encoding (BiPE)” (ICML ’24)  
  & long-document QA / summarization  
  & No  
  & Yes (inter-/intra-segment)  
  & ROUGE / accuracy  
  & Outperforms RoPE/ALiBi on 8192-token contexts in QA/summarization. \\[1ex]

\bottomrule
\end{tabular}
\caption{Length‐extrapolation methods on algorithmic tasks.}
\label{tab:related_extrap_part1}
\end{table}

%-----------------------------------------------------------------------
% Table 1b: Related Work (Part 2)
%-----------------------------------------------------------------------
\begin{table}[h!]
\centering
\small
\begin{tabular}{p{3.5cm}p{2.5cm}p{1.8cm}p{1.8cm}p{1.8cm}p{4.0cm}}
\toprule
\textbf{Citation (Year)} & \textbf{Task Type} & \textbf{Index Hints?} & \textbf{Learned Distance?} & \textbf{Metric} & \textbf{Key Extrapolation Result} \\
\midrule

McLeish et al., “Abacus Embeddings” (NeurIPS ’24)  
  & 20→100-digit addition, multiplication, sort  
  & Yes (generalized index hints; per digit index)  
  & No  
  & exact-match  
  & 97.9\% on 100-digit addition after training on 20-digit; 5× extrapolation. \\

Transformers Can Achieve Length Generalization But Not Robustly ('24)
  & 40→100-digit addition
  & Yes (per digit index)  
  & Yes (RoPE, FIRE)  
  & exact-match  
  & 99\% using index hints+FIRE RPE on 100 digit addition while trained on 40. index hints+RoPE and other PE falls to 0\%  \\
  
Duan et al., “Attention Bias Calibration (ABC)” (’24)  
  & multi-digit arithmetic (add, parity, successor)  
  & No  
  & No (Fixed attention biasing)  
  & exact-match  
  & 100\% accuracy on 50-digit addition after training on 6-digit; perfect extrapolation \(\times8\). Depends on hard-coded (unlearned) attention weights. \\[1ex]

Looped Transformers for Length Generalization (ICLR ’25)
  & arithmetic (addition)  
  & No  
  & No position encoding in general
  & exact-match  
  & Perfect only when recurrences = predetermined number of recurrent steps; degrades to 0 before and after; no autoregressive/MLM structure; supervised step count: trained on supervised shifting of input token sequence to match the correct output. \\
  
Veisi \& Mansourian, “Cable” (ArXiv ’25)  
  & language modeling (WikiText-103)  
  & No  
  & Yes (context‐aware bias)  
  & perplexity  
  & Trained at 512 tokens outperforms 1024-trained baseline at 1024 tokens. \\[1ex]

\bottomrule
\end{tabular}
\caption{Length‐extrapolation methods on language‐modeling and advanced tasks (continued).}
\label{tab:related_extrap_part2}
\end{table}

\clearpage

\section{Definitions}
\subsection{Mathematical Preliminaries}

\begin{itemize}

    \item \textbf{Relative position.}
          We view relative position as a \emph{learned
          distance function}
          $d : \{0,\dots,P-1\}\rightarrow\mathbb{R}_{\ge 0}$ that
          satisfies the axioms of a metric up to a global scale:
          $d(k,k)=0,\;\;d(k,\ell)=d(\ell,k),\;\;d(k,\ell)\le
          d(k,m)+d(m,\ell)$.
          The function is anchored at an arbitrary
          reference token at position (or time) k whose distance starts at $0$ and grows
          to a maximum of $P$ = ${T}$ at position (or time) ${k+T}$.
          During self-attention, the query and key \emph{position streams} supply sinusoidal 
          position embeddings
          $f\!\bigl(d(k,\ell)\bigr)$ whose inner product acts as a
          \emph{similarity function}; thus learning $d(\cdot)$ end-to-end is
          equivalent to learning the geometry in which similarity is
          measured. 
    
\end{itemize}
    % ------------------------------------------------------------------
\subsection{Relative Position as a Learned Metric Space}
\label{ssec:relative-metric}
% ------------------------------------------------------------------

\paragraph{Every token carries a position.}
Let the input sequence be
\[
  S \;=\; \langle x_{0},\,x_{1},\,\dots,\,x_{T-1}\rangle.
\]
To each index \(t\) we assign a non‐negative scalar coordinate
\[
  z_{t}\;\in\;[0,\,P),
\]
which serves as that token’s relative position in the sequence.  At this introductory stage, \(z_{t}\) is simply a learned value that orders token \(x_{t}\) with respect to an implicit reference point (itself learned later).  We will explain in Section~\ref{sec:histogram-update-rule-highlevel-descr} how these \(z_{t}\) arise from a stochastic update, but here they function only as latent “distance” values encoding each token’s place within the sequence.

\subsubsection{Learned distance function (idealized definition)}

\paragraph{Continuous view.}
Let \(r:[0,T]\!\to\!\mathbb{R}_{\ge 0}\) be the latent
\emph{cursor trajectory} generated by the reset/increment/decrement
gates in the limit of infinitesimal time steps.
Define its \emph{derivative/gradient}
\[
  v(t) \;=\; \tfrac{\mathrm{d}r}{\mathrm{d}t}(t)
\]
and set
\begin{equation}
    d(t,t') \;=\;
    \int_{t}^{t'} v(\tau)\,\mathrm{d}\tau,
    \quad 0 \le t < t' \le T.
    \label{eq:cont_distance}
\end{equation}
Equation~\eqref{eq:cont_distance} is the line
integral of the instantaneous jump rate; it satisfies all metric
axioms (non-negativity, identity of indiscernibles, symmetry, and the
triangle inequality) up to a global scale.

\paragraph{Discrete Definition.}
Let the discrete \emph{forward difference}
\[
  \Delta_{t} \;=\; r_{t+1} - r_{t} \;\in\; \{-1,0,+1\},
  \qquad t \in \{0,\dots,T-1\},
\]
play the role of a one‐dimensional vector field \(F(t)=\Delta_{t}\).  Because \(F\) represents a true gradient—there exists a scalar potential \(r_{t_{0}}\) such that \(F(t)=\nabla r_{t}\)—its discrete line integral over any path depends only on the end points.  In particular, for any \(0 \le t_{0} < t_{1} \le T\),
\[
  \sum_{t = t_{0}}^{\,t_{1}-1} F(t)
  \;=\;
  \sum_{t = t_{0}}^{\,t_{1}-1} \Delta_{t}
  \;=\;
  r_{t_{1}} - r_{t_{0}}.
\]
Hence the \textbf{signed displacement}
\[
  d_{\pm}(t_{0},\,t') \;=\;
  \sum_{t = t_{0}}^{\,t'-1} \Delta_{t}
  \;=\;
  r_{t'} - r_{t_{0}}
\]
is the discrete analogue of the fundamental theorem of line integrals.

\noindent In our learned metric‐space formulation, we fix a learnable reference index—without loss of generality, index \(0\), so that \(r_{0}\) is its learned coordinate.  We then identify
\[
  z_{t_{0}} \;=\; r_{t_{0}} - r_{0}
  \;=\;
  \sum_{t = 0}^{\,t_{0}-1} \Delta_{t},
\]
which ensures that each latent coordinate \(z_{t_{0}}\) exactly equals the discrete “distance” (integral of \(\Delta_{t}\)) from the reference token at index \(0\) up to index \(t_{0}\).  In other words, \(z_{t_{0}}\) is the learned integral value from the reference point through all intermediate \(\Delta_{t}\).

\section{Architectural Details}\label{sec:arc-details-start}

\subsubsection{Dynamic origin via the reset gate.}\label{sec:dynamic_origin}
We want relative distances of our model to be anchored to any arbitrary position in the sequence in a learnable fashion. The signed displacement \(d_{\pm}(t,t')=r_{t'}-r_{t}\) is correct \emph{only when the two timestamps share a common origin}—namely, the most recent position at which the reset gate fired.  
Let
\[
    \tau(t)\;=\;\max\!\bigl\{\,t_{0}\le t:\;\text{reset gate at }t_{0}=1\bigr\}
\]
denote the last reset before or at \(t\).  
We then reinterpret the cursor as a \emph{relative coordinate}
\[
    \hat{r}_{t} \;=\; r_{t} \;-\; r_{\tau(t)},
\]
so that \(\hat{r}_{\tau(t)}=0\) by definition.  
For any two points \(t < t'\), the conservative‐field argument yields
\begin{equation}
    d_{\pm}(t_{0},\,t')
    \;=\;
    \hat{r}_{t'} \;-\; \hat{r}_{t_{0}}
    \;=\;
    \sum_{t = t_{0}}^{\,t'-1} \Delta_{t},
    \label{eq:dynamic_origin}
\end{equation}
but \eqref{eq:dynamic_origin} is meaningful \emph{only if} \(\tau(t_{0})=\tau(t')\) (i.e.\ both points are measured relative to the same reset).  Whenever \(\tau(t_{0})\neq\tau(t')\)—for example, if a reset occurs between \(t_{0}\) and \(t'\)—the model first reanchors the origin at \(r_{\tau(t')}\), yielding a piece‐wise path integral whose segments are each evaluated with respect to their local origin. Refer to~\ref{sec:dynamic-origin-other-interpretations} for reasoning behind this decision and interpretations.

\subsubsection{Learned path–integral distance \textit{distribution}.}
In practice, rather than committing to a single shortest‐path distance,  
our model treats the latent cursor trajectory  
\(
\mathcal{Z}_{k\!\to\!\ell}
    =\bigl(z_{k},z_{k+1},\dots,z_{\ell}\bigr)
\)
as a \emph{stochastic path} whose increments  
\(
\Delta_{t}=z_{t+1}-z_{t}\in\{-1,0,+1\}
\)
are drawn from the Bernoulli gates described in
Section~\ref{sec:histogram-update-rule-highlevel-descr}.   
The \emph{signed displacement}
\(
D_{k\!\to\!\ell}
    =\sum_{t=k}^{\ell-1}\!\Delta_{t}
\)
is therefore a \emph{path integral under uncertainty}.  
Formally, the probability mass at offset \(d\) is the Feynman–Kac–type
functional integral over all admissible paths.\\

\noindent Its \emph{probability mass function} marginalizes over \emph{all} paths
of elementary shifts
Let
$\bm{\Delta}_{k:\ell-1}
  = (\Delta_{k},\Delta_{k+1},\dots,\Delta_{\ell-1})
  \in \{-1,0,+1\}^{\,\ell-k}$
be the sequence of elementary shifts.
Then

\begin{equation} \label{eq:pmf_no_reset}
\Pr\bigl(D_{k\!\to\!\ell}=d\bigr)
\;=\;
\sum_{\bm{\Delta}_{k:\ell-1}\in\{-1,0,+1\}^{\ell-k}}
\Bigl[
    \prod_{t=k}^{\ell-1}
        p\bigl(\Delta_{t}\mid\theta_t\bigr)
\Bigr]\,
\mathbf{1}\!\Bigl\{
   \sum_{t=k}^{\ell-1}\Delta_{t}=d
\Bigr\}.
\end{equation}

\smallskip

\medskip
\noindent\textbf{Refinement with a dynamic origin.} 
Whenever the reset gate (Section~\ref{sec:dynamic_origin}) "fires", it re-anchors the coordinate frame; any increments before that
time no longer contribute to the net displacement.
Introduce the reset indicator \(R_{t}\in\{0,1\}\), so that
the effective displacement from \(k\) to \(\ell\) is governed by

\begin{equation}\label{eq:pmf_with_reset}
\Pr\bigl(D^{\star}_{k\!\to\!\ell}=d\bigr)
=
\sum_{\substack{
    \bm{R}_{k:\ell-1}\in\{0,1\}^{\ell-k}\\[-2pt]
    \bm{\Delta}_{k:\ell-1}\in\{-1,0,+1\}^{\ell-k}
}}
\Bigl[\,
  \prod_{t=k}^{\ell-1}
    p\bigl(R_{t}\mid\theta_t\bigr)\,
    p\bigl(\Delta_{t}\mid\theta_t\bigr)
\Bigr]
\;\chi_{k,\ell}(\bm R,\bm\Delta;d).
\end{equation}

\noindent Here we have introduced the \emph{path-indicator} function
\(\chi_{k,\ell}\), which enforces that only the increments after
the last reset contribute:

\begin{equation}\label{eq:path_indicator}
\chi_{k,\ell}(\bm R,\bm\Delta;d)
\;:=\;
\mathbf{1}\Bigl\{
  \sum_{t=k}^{\ell-1}
    \Delta_{t}\,
    \prod_{u=t+1}^{\ell-1}(1 - R_u)
  = d
\Bigr\}.
\end{equation}

%------------------------------------------------------------------

%---------------------------------------------------------------

\noindent Equation~\eqref{eq:pmf_with_reset} reduces to
\eqref{eq:pmf_no_reset} when \(R_{t}\!=\!0\;\forall\,t\).
Intuitively, each reset \emph{excises} the history before it, so the
path integral accumulates only those steps measured in the current
relative frame.  
This refinement aligns the probabilistic distance with the geometric
definition of a dynamic origin introduced in
Eq.~\eqref{eq:dynamic_origin}.\\

\noindent Here each block $[s_{j},s_{j+1})$ contains \emph{no} intervening reset,
so the local displacement distribution inside the product is exactly
the original Feynman–Kac sum.
The overall distance distribution is therefore a \emph{mixture} of
block-wise integrals glued together at reset landmarks—mirroring how
the model dynamically re-anchors its coordinate system whenever the
reset gate fires.

\subsubsection{Position Update with Copy Branch}
\label{sec:def-copy-branch}

We generalize the classic reset operation (which re-centers the relative position to a fixed origin, typically 0) by allowing a “copy” operation: at any time $t$, the model may *copy* its reference position from any previous position, rather than only resetting to the start. This defines a learned “constant of integration” for the position histogram update. Formally, the position update becomes:
\[
\bm{h}_{t+1} = \mathcal{U}\left(\bm{h}_t; \mathbf{p}_{\text{reset}}, \mathbf{p}_{\text{incr}}, \mathbf{p}_{\text{decr}}, \mathbf{p}_{\text{keep}}, \mathbf{p}_{\text{copy}}\right)
\]
where $\mathbf{p}_{\text{copy}}$ is a (possibly sparse) distribution over copy targets and a null option.

\paragraph{Implementation}
\label{sec:arch-copy-branch}

In our architecture we add a \textbf{copy branch} to the position histogram update for a subset learned learned relative position cursors, corresponding 1:1 to a subset of the query vectors in standard dot product attention. This branch enables the model, at each step, to optionally copy its reference position from any previous key (or token) position, not just reset to zero. This operation is implemented by augmenting the stochastic update matrix with a copy distribution $\mathbf{p}_{\text{copy}} \in \Delta^{P+1}$ over all candidate positions and a null (“no-copy”) option. Importantly, for the experiments in this paper, we only apply the copy branch for a subset of the tasks. Refer to Experiments (\ref{sec:experiments}) for more details.

\subsubsection{Position Update with Copy Branch}
\label{sec:def-copy-branch}

We generalize the classic reset‐only update (which always re‐anchors to position 0) by adding a learned “copy” branch: at time \(t\), the model may copy its reference position from any previous index \(k\), instead of only resetting to 0. Concretely, let
\[
\bm{h}_t \;\in\;\Delta^{P-1}
\]
be the categorical distribution over all \(P\) possible positions at step \(t\), and let
\[
\mathbf{p}_{\text{copy}}(t)\;\in\;\Delta^{P}
\]
be a (learned) distribution over “copy‐from index” \(k=0,\dots,P-1\) plus a terminal “no‐copy” slot.  Define
\[
p_{\text{no‐copy}}(t) \;=\; 1 \;-\;\sum_{k=0}^{P-1}p_{\text{copy}}(t,k)\,.
\]
Then the next‐step histogram becomes a mixture of (1) the standard reset/​incr/​decr update—only if we do not copy—and (2) direct jumps into one‐hot modes at each \(k\) if we do copy:

\[
\begin{aligned}
\bm{h}_{t+1}
\;=\;
p_{\text{no‐copy}}(t)\;\mathcal{U}_{\text{standard}}\bigl(\bm{h}_t\bigr)
\;+\;
\sum_{k=0}^{P-1} p_{\text{copy}}(t,k)\;\bm{v}_k,
\end{aligned}
\tag{3.8$^{\prime}$}
\]
where:
\begin{itemize}
  \item \(\mathcal{U}_{\text{standard}}(\bm{h}_t)\in\Delta^{P-1}\) is the usual \emph{reset/incr/decr/keep} update applied to \(\bm{h}_t\) (i.e.\ exactly as in Eq.~(3.7) with no copy branch).  
  \item \(\bm{v}_k\) denotes the one‐hot vector whose only nonzero entry is at position \(k\).  
  \item \(p_{\text{no‐copy}}(t) = 1 - \sum_{k=0}^{P-1}p_{\text{copy}}(t,k)\) ensures that \(\sum_{i=0}^{P-1}[\bm{h}_{t+1}]_i=1\).  
\end{itemize}
\noindent In other words, with probability \(p_{\text{no‐copy}}(t)\) we proceed exactly as in the standard histogram update, and with probability \(p_{\text{copy}}(t,k)\) we force the next histogram to collapse entirely to position \(k\).  This minimal modification recovers the usual reset‐only behavior when \(\mathbf{p}_{\text{copy}}(t)\) is a delta at the “null” slot, but allows flexible copying from any past index when desired.  See Algorithm 3 for the fully detailed implementation. In practice, \(p_{\text{copy}}(t,k)\) is determined using a standard, causally masked self-attention matrix.\\

\noindent This generalizes the reset operation ($\mathbf{p}_{\text{copy}}$ as a delta at 0) and allows flexible reference anchoring—crucial for dynamic tasks such as copying from arbitrary indices. See Alg.~\ref{alg:histogram} for details.

\subsubsection{Recurrent Neural Network Architecture }
To compute for latents $\theta_{t}$ described in eq.~\ref{eq:pmf_with_reset}, we implement the gating mechanism via a Gated Recurrent Unit (GRU) with a hidden state dimension of 100, whose weights are learned end-to-end alongside the rest of the model. While our architecture does not require a recurrent model to extrapolate on many (if not most) tasks, we have included it in order to reduce the number of transformer layers and to explore the usefulness of a more recurrent architecture. Furthermore, while the Histogram Update Rule is also computed sequentially, the implementation of the GRU, as well the Histogram Update Rule can be significantly parallelized in terms of overhead via kernel fusion, which we leave for future work.

\subsection{Histogram Update Rule}\label{sec:histogram-update-rule-highlevel-descr}
In Section \ref{sec:arc-details-start}, we outlined the basic histogram update rule mathematically. Here we summarize the histogram update algorithm\footnote{excluding the 'copy branch' initially for simplicity}. The \textit{optimized} algorithm in pseudocode, including the 'copy branch', is detailed in Algorithm~\ref{alg:histogram} of Appendix.\\

\noindent We then briefly analyze the computational complexity of a naive update rule implementation, followed by the complexity of our optimized approach. Note that the copy branch of the full algorithm in Appendix~\ref{alg:histogram} does not increase the optimal upper bound in the computational complexity of our algorithm.\\

\noindent\textbf{Step 1: Reset Branch.} \\
If reset occurs, all mass flows to position $0$, then is split by action:
\begin{align*}
    \text{Total mass:} \quad & S = \sum_{i=0}^{P-1} [\bm{h}_t]_i \\
    \text{Reset increments:}   &\quad r_{\mathrm{inc}} = S \cdot p_{\mathrm{reset}}(t) \cdot p_{\mathrm{incr}}(t) \\
    \text{Reset decrements:}   &\quad r_{\mathrm{dec}} = S \cdot p_{\mathrm{reset}}(t) \cdot p_{\mathrm{decr}}(t) \\
    \text{Reset keep:}         &\quad r_{\mathrm{keep}} = S \cdot p_{\mathrm{reset}}(t) \cdot p_{\mathrm{keep}}(t)
\end{align*}
The reset-branch histogram is
\[
[\bm{h}_{t+1}^{\mathrm{(reset)}}]_i =
\begin{cases}
    r_{\mathrm{inc}}, & i=1 \\
    r_{\mathrm{keep}} + r_{\mathrm{dec}}, & i=0 \\
    0, & \text{else}
\end{cases}
\]

\noindent\textbf{Step 2: No-Reset Branch.} \\
If no reset, the histogram is shifted left/right/stays according to the action probabilities:
\begin{align*}
[\bm{h}_{t+1}^{\mathrm{(no-reset)}}]_i =\;&
[\bm{h}_t]_i     \cdot (1 - p_{\mathrm{reset}}(t)) \cdot p_{\mathrm{keep}}(t) \\
&+ [\bm{h}_t]_{i-1} \cdot (1 - p_{\mathrm{reset}}(t)) \cdot p_{\mathrm{incr}}(t) \\
&+ [\bm{h}_t]_{i+1} \cdot (1 - p_{\mathrm{reset}}(t)) \cdot p_{\mathrm{decr}}(t)
\end{align*}
(Assume $[\bm{h}_t]_{-1} = [\bm{h}_t]_P = 0$.)\\

\noindent\textbf{Step 3: Sum and Sharpen using the $\gamma$ parameter for each cursor $c$.}
\[
\bm{h}_{t+1} = \bm{h}_{t+1}^{\mathrm{(reset)}} + \bm{h}_{t+1}^{\mathrm{(no-reset)}}
\]
In order to prevent oversmoothing \label{sec:gamma-sharpen} we use the power-sharpening (temperature $\gamma > 1$), then\footnote{See Appendix~\ref{app:power-sharpen} for full derivation and ablations.}:
\[
\bm{h}_{t+1}^{\mathrm{(sharp)}}[i] = \frac{\left([\bm{h}_{t+1}]_i + \varepsilon\right)^\gamma}{\sum_j \left([\bm{h}_{t+1}]_j + \varepsilon\right)^\gamma}
\]

\subsubsection{Computational Complexity}\label{sec:computational-complexity}

A naïve way to model the latent position after $t$ steps is to enumerate
\emph{all} action sequences of length~$t$.  
Because reset is binary ($\text{yes}/\text{no}$) and the action gate\footnote{omitting the copy branch for simplicity} is ternary
($\text{inc},\text{dec},\text{keep}$), the joint number of branches grows as
\[
\underbrace{(2 \times 3)}_{\text{reset}\,\times\,\text{action}}^{t\ \text{times}}
\;=\; 6^{\,t},
\]
i.e.\ exponential in~$t$.  
Maintaining an explicit categorical distribution over those paths would be
prohibitive.

% \noindent\textbf{Efficient evaluation.}  
\paragraph{Efficient Evaluation}
% A naïve enumeration of all reset/increment/decrement paths between two
% tokens \(x_{k}\) and \(x_{\ell}\) grows as
% \(\mathcal{O}(6^{\,\ell-k})\).
Our algorithm avoids this blow-up by marginalizing \emph{incrementally}:
The stochastic matrix update below (\ref{sec:stochastic-matrix-form}) updates a length-\(P\) histogram
\(\bm{h}_{t}\) in \(\mathcal{O}(P)\) time for each \emph{adjacent}
step \(t\!\to\!t{+}1\).
Because the model is Markovian given we know \(\theta_{t}\), \(\bm{h}_{t}\) already contains the
\emph{exact} marginal over \emph{all} paths that end at step~\(t\).
Therefore, for any \(k<\ell\) we can reconstruct the
signed-displacement distribution solely from
those two histograms via a discrete cross-correlation:
\begin{equation}
    \Pr\!\bigl(D_{k\!\to\!\ell}=d\bigr)
    \;=\;
    \sum_{i=0}^{P-1}
       \bm{h}_{k}[i]\;
       \bm{h}_{\ell}\!\bigl[i+d\bigr],
    \qquad
    -(P{-}1)\le d\le P{-}1.
    \label{eq:rel-dist-pmf-path}
\end{equation}
Intuitively, \(\bm{h}_{k}[i]\) gives the probability that the cursor is
at position \(i\) after all paths up to step~\(k\);  
\(\bm{h}_{\ell}[i{+}d]\) gives the probability of being at
\(i{+}d\) at step~\(\ell\).
Multiplying and summing over \(i\) integrates out the latent positions
while enforcing the net displacement constraint,
exactly matching the Feynman–Kac sum in
Eq.~\eqref{eq:pmf_with_reset}.
Thus the total cost remains \(\mathcal{O}(P\,T)\) for the entire
sequence.

\subsubsection{Stochastic Matrix Formulation}\label{sec:stochastic-matrix-form}

The update can be compactly expressed as:
\[
\bm{h}_{t+1} = \bm{h}_t^\top \mathbf{M}_t
\]
where $\mathbf{M}_t \in \mathbb{R}^{P \times P}$ is a row-stochastic matrix with entries determined by the above formulas.

\paragraph{Block-structured transition matrix.}
Our update circumvents the blow-up by
\emph{marginalizing} over the branching decisions in closed form.  The effect of
every reset/action pair can be written as a sparse, band-diagonal block in a
$P{\times}P$ stochastic matrix~$\mathbf{M}_t$:
\[
\bm{h}_{t+1} \;=\; \bm{h}_t^\top \mathbf{M}_t,
\qquad
\mathbf{M}_t
\,=\, 
\mathbf{M}_{t}^{\text{reset}}
\;+\;
\mathbf{M}_{t}^{\text{no-reset}} ,
\]
where each block has only
$\mathcal{O}(P)$ non-zero entries (three per row after clamping).
Because we apply $\mathbf{M}_t$ implicitly with simple
\texttt{scatter\_add} operations, the per-step cost is
\[
\boxed{\mathcal{O}(P)} ,
\]
rather than the $\mathcal{O}(P^2)$ dense matrix–vector multiply or the
$\mathcal{O}(6^{t})$ path enumeration.
The full pseudocode for this update rule and matrix formulation is described in Alg.~\ref{alg:histogram}.

\paragraph{Interpretation.}
Thus the histogram acts as a \emph{compressed} representation of the full
branching Bernoulli–categorical process:
all $6^{t}$ trajectories are still present, but their probability mass is pooled by position.
This provides an exact, differentiable update whose complexity scales
linearly with the support size~$P$ and the sequence length~$T$
($\mathcal{O}(P\,T)$ total), enabling practical training and inference on long sequences.
This block-structure corresponds to a
\emph{non-stationary HMM}; its conditional independence assumptions
also let us cast the model as a first-order linear-chain CRF
conditioned on the input tokens. However, this connection to probabilistic graphical models is only drawn for intuition. We take the linear (convex) combination of possible "positions", whose coefficients are represented by ${h}_{t}$, via \textit{superposition}~\cite{elhage2022toymodels} of features, i.e. \textit{position stream} vectors. See~\ref{convex-combinations-instead-of-probabilistic-mixture} for details.

\paragraph{Complexity of the "copy branch".} Finally, with the addition of the copy branch defined in ~\ref{sec:def-copy-branch}, the computation complexity remains unchanged. At each time step \(t\), computing the "copy branch" distribution \(\mathbf{p}_{\text{copy}}(t)\) via a single self‐attention query against all previous keys requires one dot‐product of the query vector with each of the \(t\) key vectors, followed by a softmax over \(t\) scores-i.e.\ \(O(t)=O(T)\) work.  Once \(\mathbf{p}_{\text{copy}}(t)\) is available, forming the mixture in Eq.~(3.8) involves (1) applying \(\mathcal{U}_{\text{standard}}(\bm{h}_t)\), which itself is \(O(P)\), scaling by \(p_{\text{no‐copy}}(t)\), and (2) accumulating \(\sum_{k=0}^{P-1}p_{\text{copy}}(t,k)\,\bm{v}_k\), which is also \(O(P)\).  Therefore, the total per‐step cost is
\[
  O\bigl(t + P\bigr)\;=\;O\bigl(\max\{P,\,T\}\bigr).
\]
In practice one takes \(P=O(T)\), so each histogram update remains \(O(T)\) per time step.

\subsection{From distance to \textit{similarity}.}
Given a metric $d\colon\!\{0,\dots,P{-}1\}^{2}\!\to\!\mathbb{R}_{\ge 0}$
we define a \emph{similarity measure}
\begin{equation}
    \sigma(k,\ell)
    \;=\;
    \kappa\!\bigl(d(k,\ell)\bigr),
    \qquad
    0\le k,\ell<P ,
    \label{eq:similarity}
\end{equation}
where $\kappa\!:\mathbb{R}_{\ge 0}\!\to\![0,1]$ is a
monotonically–decreasing kernel that satisfies
$\kappa(0)=1$ and $\kappa(r_1)\!>\!\kappa(r_2)$ whenever
$r_1<r_2$.
Common choices include

\smallskip
\begin{tabular}{@{}ll@{}}
\textbf{Gaussian (RBF):} &
$\displaystyle
  \kappa_{\text{RBF}}(r)=\exp\!\bigl(-\tfrac{r^{2}}{2\lambda^{2}}\bigr)$
\\[6pt]
\textbf{Laplacian:} &
$\displaystyle
  \kappa_{\text{Lap}}(r)=\exp\!\bigl(-\tfrac{r}{\lambda}\bigr)$
\\[6pt]
\textbf{Cosine–sinusoid (used here):} &
$\displaystyle
  \kappa_{\text{cos}}(r)=
      \frac{1}{d/2}\sum_{m=0}^{d/2-1}\!
      \cos\!\bigl(\tfrac{r}{10000^{\,2m/d}}\bigr)$ ,
\end{tabular}

\smallskip\noindent
with bandwidth~$\lambda$ or, in the sinusoidal case,
a log–frequency grid identical to Transformer‐style position
encodings.  
When the encoding map $f(\cdot)$ of
Section~\ref{sec:similarity-via-positional`w-streams} satisfies nearest-neighbor properties,
the inner product
$\langle f(k),f(\ell)\rangle$ realizes~\eqref{eq:similarity} exactly,
so the attention mechanism respects the metric induced by~$d(k,\ell)$.

\paragraph{Similarity via positional \emph{streams}.}
We use periodic functions to define our similarity function
implementation, as they give embedding vectors that can be linearly combined using the probability distribution \(\bm{h_{t}}\). In particular, we follow the original Transformer:\footnote{\texttt{div\_term} in the reference
\textsc{PyTorch}/\textsc{NumPy} routine is
$\exp\!\bigl(-\ln 10000 \cdot 2i/d_{\mathrm{pe}}\bigr)$.}
let $d_{\mathrm{pe}}$ be the block width and
\[
  f_{2i}(z)=\sin\!\Bigl(z\,e^{-\ln 10000\cdot 2i/d_{\mathrm{pe}}}\Bigr),\qquad
  f_{2i+1}(z)=\cos\!\Bigl(z\,e^{-\ln 10000\cdot 2i/d_{\mathrm{pe}}}\Bigr),
  \qquad i=0,\dots,\frac{d_{\mathrm{pe}}}{2}-1 .
\]

\noindent\textbf{Position stream superposition.}  
Instead of collapsing the histogram to a single expected coordinate,
we form the \emph{position stream}
\(
\bm{e}_{t}=\sum_{k=0}^{P-1}\bm{h}_{t}[k]\;f(k),
\)
so that \emph{every} plausible position contributes its full encoding.

\noindent\textbf{Hierarchical PE.}  
\label{sec:hierarchical-pe}
We concatenate four such superposition features evaluated at the \emph{same}
position to obtain a length-$4d_{\mathrm{pe}}$ embedding
\(
  g(t)=\bigl[\bm{e}^{\,0}_{t}\,\|\,\bm{e}^{\,1}_{t}\,\|\,\bm{e}^{\,2}_{t}\,\|\,\bm{e}^{\,3}_{t}\bigr],
\)
or, more generally,
\[
  g(t)
  \;=\;
  \bigl[
      \bm{e}^{\,0}_{t}
      \,\|\,
      \bm{e}^{\,1}_{t}
      \,\|\,
      \dots
      \,\|\,
      \bm{e}^{\,C-1}_{t}
  \bigr]
  \;\in\;\mathbb{R}^{C\,d_{\mathrm{pe}}},
\qquad
  \text{where}\;\;
  \bm{e}^{\,c}_{t}
  \;=\;
  \sum_{k=0}^{P-1} \bm{h^{c}}_{t}[k]\;f(k),
  \quad
  c=0,\dots,C-1 .
\]

which allows the model to learn coarse-to-fine spatial features. For the detailed algorithm, refer to Alg.~\ref{alg:superposition}.

% \noindent\textbf{Learned similarity.}  
% During self-attention the query and key streams interact through
% % \[
% \begin{equation}
%   \alpha_{h,c}\,
%   \bigl\langle\bm{g}^{(q,c)}_{t},\,\bm{g}^{(k,c)}_{s}\bigr\rangle,
% % \]
% \end{equation}\label{sec:alpha-coeff}
% where \({s \geq t}\), $\alpha_{h,c}\!\geq\!0$ is a \emph{learnable,
% head-specific} scaling coefficient.
% Because\;
% $\langle g(k),g(\ell)\rangle\propto\sum_{j=0}^{3}\cos\!\bigl(\omega_{1...{d_{pe}/2}}(k^{j}-\ell^{j})\bigr)$,
% the similarity decays smoothly with the
% distance $d(k,\ell)=|k-\ell|$, yet exhibits a sharp,
% Gaussian-like peak at $k=\ell$ once the sum over multiple frequencies
% is taken—providing local precision and long-range smoothness that are
% crucial for reliable length extrapolation.

% Notably, the sinusoidal superposition encoding
% \(
% \bm{e}_{t}=\sum_{j}\bm{h}_{t}[j]\,f(j)
% \)
% maps \emph{every coefficient of the histogram} into phase information;
% subsequent dot‐product attention therefore receives a signal that
% implicitly contains \emph{all} moments of
% $D_{k\!\to\!\ell}$, not just its mean.

\noindent\textbf{Learned similarity.}  
During self-attention the query and key streams interact through
\begin{equation}
  \alpha_{h,c}\,
  \bigl\langle\bm{g}^{(q,c)}_{t},\,\bm{g}^{(k,c)}_{s}\bigr\rangle,
  \label{eq:alpha-coeff}
\end{equation}
where \({s \geq t}\), $\alpha_{h,c}\!\geq\!0$ is a \emph{learnable,
head-specific} scaling coefficient.
Because\;
$\langle g(k),g(\ell)\rangle\propto\sum_{j=0}^{3}\cos\!\bigl(\omega_{1\dots d_{pe}/2}(k^{j}-\ell^{j})\bigr)$,
the similarity decays smoothly with the
distance $d(k,\ell)=|k-\ell|$, yet exhibits a sharp,
Gaussian-like peak at $k=\ell$ once the sum over multiple frequencies
is taken—providing local precision and long-range smoothness that are
crucial for reliable length extrapolation.
For the detailed algorithms, refer to Alg.~\ref{alg:abs‐ln‐optim} and Alg.~\ref{alg:compute-position-scores}.

Notably, the sinusoidal superposition encoding
\(
\bm{e}_{t}=\sum_{j}\bm{h}_{t}[j]\,f(j)
\)
maps \emph{every coefficient of the histogram} into phase information;
subsequent dot-product attention therefore receives a signal that
implicitly contains \emph{all} moments of
$D_{k\!\to\!\ell}$, not just its mean.

\paragraph{Hybrid unnormalised score.}
For the hierarchical positional-encoding variant we mix the standard
dot-product term with the learned similarity term to obtain the
head-wise unnormalised self-attention score
\label{sec:alpha-coeff}
\begin{equation}
  \tilde{a}^{(h)}_{t,s}
  \;=\;
  \mu^{(h)}
  \bigl\langle \bm{q}^{(h)}_{t},\,\bm{k}^{(h)}_{s} \bigr\rangle
  \;+\;
  \bigl(1-\mu^{(h)}\bigr)
  \,\alpha_{h,c}\,
  \bigl\langle\bm{g}^{(q,c)}_{t},\,\bm{g}^{(k,c)}_{s}\bigr\rangle,
  \qquad \mu^{(h)}\in[0,1],
  \label{eq:hybrid-score}
\end{equation}

where \(\mu\) (learned per head) gates between the conventional
self-attention score and the frequency-based learned similarity from
Section 5.2.  Setting \(\mu=1\) recovers vanilla self-attention, whereas
\(\mu=0\) relies purely on the hierarchical positional signal; learning
\(\mu\) allows the model to interpolate continuously between the two.

\section{Experiments}
\label{sec:exp-setup}

\paragraph{Task suite overview.}
Table~\ref{tab:tasks-1} - Table~\ref{tab:tasks-2} summarize the eight self-supervised problems that constitute our
benchmark.  All of them share one crucial property: \textbf{only the tokens to the right of the “=” sign are supervised,
so the model must learn to read the left side and predict the completion}. We only measure exact-match accuracy; therefore, all tokens, including chain-of-thought tokens, must be predicted correctly in order for the answer to be marked correct. Furthermore, we require all end of sequence (EOS) tokens to be outputted in order for the model's answer to be marked correct.
The tasks span arithmetic (Addition, Multiplication), list processing
(Reverse, Odds First), stack semantics, program-like navigation
(\textsc{SCAN-CoT}), and two forms of string copying, thereby exercising both
symbolic and positional generalization.\\
\\
\paragraph{SCAN-CoT expansion.}
Because \textbf{SCAN-CoT} is derived 1:1 from a fixed-dataset of only $\sim$20k unique examples via chain-of-thought tokens, we visualize its length distribution in Appendix~\ref{sec:scan-len-dataset-distr}. For stratifying results, we measure accuracies on the entire test set, All train/test examples are derived 1:1 with the originals by expanding the correct output tokens via the chain-of-thought tokens described in Table~\ref{tab:tasks-2}.

\paragraph{Multiplication CoT expansion.}
In our chain‐of‐thought representation of $a\times b$, we unfold the standard grade‐school long‐multiplication via the distributive law:
\[
  a \times b \;=\;
  a \times \Bigl[\sum_{i} d_i \times 10^{p_i}\Bigr]
  \;=\;
  \sum_{i}\bigl(a \times d_i\bigr)\times 10^{p_i}.
\]
Here each digit $d_i$ of the multiplier $b$ produces a partial product $a\times d_i$, and is then shifted by $p_i$ places (i.e.\ padded with $p_i$ zeros) according to its place value.  We annotate each term as
\[
  d_i\,\bigl(d_i\!\to p_i\bigr)
\]
to indicate that the partial product $a\times d_i$ is followed by $p_i$ zeros.  This sequence of weighted, place‐shifted multiplications fully encodes the long‐multiplication algorithm, without ever performing the final addition. The correct output tokens via chain-of-thought is described in Table~\ref{tab:tasks-2}.

\subsubsection{Copy–branch ablation.}
Our histogram encoder can optionally perform a 'copy branch' jump
(§ \ref{sec:arch-copy-branch}) via a dedicated branch.  We toggled this feature
per experiment as listed below:
\paragraph{For \textbf{Addition, Odds First, Stack Manipulation, and CoT-only Multiplication}}
the copy branch was \emph{disabled} on every head.
\paragraph{For \textbf{Reverse}, \textbf{SCAN-CoT}, and \textbf{dyn\_str\_cpy}}, the \textbf{copy branch} was \emph{enabled} on a small
subset—one out of every five query cursors. For the corresponding key cursors, from which query could perform a copy branch operation, the corresponding key cursor histogram was fixed to increment-only, so that the relative positions corresponded to exactly to the corresponding tokens' \textbf{absolute} instead of 
learnable relative positions. We emphasize the generality of this approach to other text retrieval tasks in practice, because for a small fraction of our position encodings, it is useful to allow our model to make them absolute rather than relative positions, so as to give each token a \textit{unique} position encoding. Therefore, some of our query vectors are \textit{guaranteed} to be able to attend to each past token uniquely.

\noindent Empirically, with the exception of \textbf{SCAN and dyn\_str\_cpy} we observed no extrapolation difference when that key cursor was allowed to train, but we keep the setting fixed for consistency.  All other experiments also ran with the copy branch turned off.

%---------------------------------------------------------------
%  Table: Experimental tasks (completion-side supervision only)
%  Requires: \usepackage{tabularx,booktabs}
%---------------------------------------------------------------

\begin{table}[H]
  \caption{Self‐supervised tasks used in our experiments.  
           Only the tokens \textbf{to the right of the “=” sign} are optimized.}
  \label{tab:tasks-1}
  \small
  \centering
  \begin{tabularx}{\linewidth}{@{}l X X X@{}}
    \toprule
    \textbf{Task} &
    \textbf{Input prefix (left of ``='')} &
    \textbf{Target completion (right of ``='' )} &
    \textbf{Example} \\
    \midrule
    % -----------------------------------------------------------------
    %  Replace ONLY the “Addition” row with the block below
    % -----------------------------------------------------------------
    Addition\label{task:cot-addition} &
    Two integers written least-significant-digit first, separated by “+”, then “=”. &
    Comma-separated chain-of-thought; each step emits the four per-digit tokens  
    $\langle d_a,d_b,c_{\text{out}},d\rangle$
    (​addend\,a digit, addend\,b digit, outgoing carry, result digit).
    After the list we write “$\rightarrow$” followed by the reversed
    sum and a final “.”.  
    The answer is marked correct only when **all** CoT steps
    \emph{and} the final digits are correct. &
    \begin{minipage}[t]{0.78\linewidth}\scriptsize\ttfamily
    8 2 9 + 0 3 = 8 0 0 8 , 2 3 0 5 , 9 0 0 9 → 8 5 9 .
    \end{minipage}\\[2pt]

    Copy &
    Random digit sequence. &
    Identical sequence. &
    \texttt{8349216=8349216.} \\[2pt]

    Reverse &
    Random digit sequence. &
    Sequence reversed. &
    \texttt{8349216=6129438.} \\[2pt]

    Odds First &
    Length-$L$ digit sequence over $\{0,\dots,V-1\}$. &
    Digits at odd indices (1-based) concatenated in order, then the even-indexed digits. &
    \texttt{012345=135024.} \\[2pt]

    % --------------------------------------------------------------
    %  Replace ONLY the Stack-Manipulation row with the block below
    % --------------------------------------------------------------
    Stack Manipulation &
    Initial binary stack ($0/1$ written bottom$\to$top) followed by actions
    \{2: POP, 3: PUSH 0, 4: PUSH 1\}; total length~$\ell$, then “=”. &
    Resulting stack written \emph{top-to-bottom}, a terminator “2”, and
    zero-padding to reach length $\ell+1$. &
    \texttt{0 1 1 0 4 2 2 = 1 1 0 2 0 0 0 0.}\, \\[2pt]

  \end{tabularx}
\end{table}

\begin{table}[H]  % ← change to [H] if you want it exactly “here”

  \caption{Self-supervised tasks used in our experiments (Continued).  
           Only the tokens \textbf{to the right of the “=” sign} are optimized.}
  \label{tab:tasks-2}
  \small
  \centering
  \begin{tabularx}{\linewidth}{@{}l X X X@{}}
    \toprule
    \textbf{Task} &
    \textbf{Input prefix (left of “=”)} &
    \textbf{Target completion (right of “=”)} &
    \textbf{Example} \\
    \midrule
% -----------------------------------------------------------------
%  ⬇︎  replace the old SCAN-CoT row with this one  ⬇︎
% -----------------------------------------------------------------
% --------------------------------------------------------------
%  Replace ONLY the SCAN-CoT row with the block below
% --------------------------------------------------------------
\addlinespace[4pt]
SCAN-CoT\footnotemark &
All train/test data are obtained by converting the full
\emph{length-split} split of SCAN into chain-of-thought form. &
Chain-of-thought for every sub-clause, followed by “$\rightarrow$”, then
\emph{all} low-level action tokens.  The model must predict the full
CoT, not just the final actions. &
\begin{minipage}[t]{0.78\linewidth}
  \scriptsize\ttfamily\raggedright
  walk around left thrice and look opposite right = walk around left +
  walk around left + walk around left and look opposite right
  $\rightarrow$ walk around left : I\_TURN\_LEFT I\_WALK I\_TURN\_LEFT
  I\_WALK I\_TURN\_LEFT I\_WALK I\_TURN\_LEFT I\_WALK walk around left :
  I\_TURN\_LEFT I\_WALK I\_TURN\_LEFT I\_WALK I\_TURN\_LEFT I\_WALK
  I\_TURN\_LEFT I\_WALK walk around left : I\_TURN\_LEFT I\_WALK
  I\_TURN\_LEFT I\_WALK I\_TURN\_LEFT I\_WALK I\_TURN\_LEFT I\_WALK
  look opposite right : I\_TURN\_RIGHT I\_TUR​N\_RIGHT I\_LOOK .
\end{minipage}\\[2pt]

% -----------------------------------------------------------------

    dyn\_str\_cpy &
    Pair “\(s,d\)” where \(d\) occurs exactly once in \(s\). &
    Substring of \(s\) starting at that unique \(d\), where the start index 
    \[
    \begin{aligned}
      t &\sim \mathrm{Uniform}\bigl(\{1,\dots,n\}\bigr),\\
      n &= \lvert s\rvert
    \end{aligned}
    \]
    so output \(s_{t:}\). &
    \texttt{5839472,3=39472.} \\[2pt]

    Integer‐Multiplication (CoT only) &
      “$a\!\times b$=”. &
      Program‐like CoT string that rewrites $a\times b$ as
      $a\times[\,$\emph{weighted digit mappings}$]$; the final product is \emph{not} required. &
      \begin{minipage}[t]{\hsize}\ttfamily\small
        675x1259=
        675x[1*(1→1)(2→0)(5→0)(9→0)\\
        \qquad\;+2*(2→1)(5→0)(9→0)\\
        \qquad\;+5*(5→1)(9→0)\\
        \qquad\;+9*(9→1)].
      \end{minipage} \\
    \bottomrule
  \end{tabularx}
\end{table}

% now place the footnotetext outside (and after) the table
\footnotetext{%
  Adapted directly from the SCAN~\cite{lake2018} \emph{length-split} corpus
  (\nolinkurl{https://github.com/brendenlake/SCAN/tree/master/length_split});
  the original command is expanded into two supervised chain-of-thought steps.%
}

\pagebreak

\subsection{Evaluation Protocol}
\begin{figure}[H]
  \centering
  \includegraphics[width=0.8\linewidth]{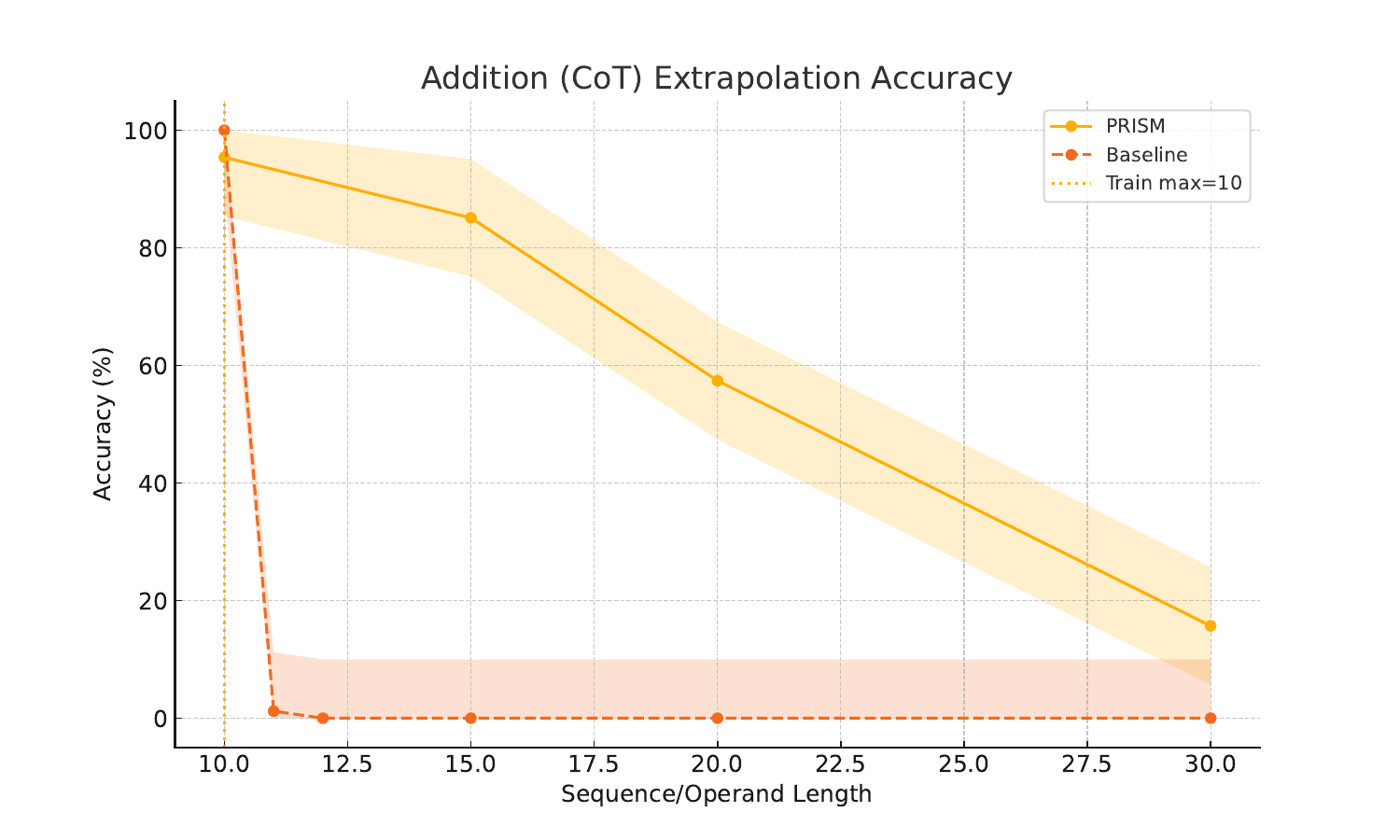}
  \caption{Addition (CoT) extrapolation accuracy. PRISM vs.\ Baseline. Dashed line marks the training max length ($10$).}
  \label{fig:addition}
\end{figure}

\begin{figure}[H]
  \centering
  \includegraphics[width=0.8\linewidth]{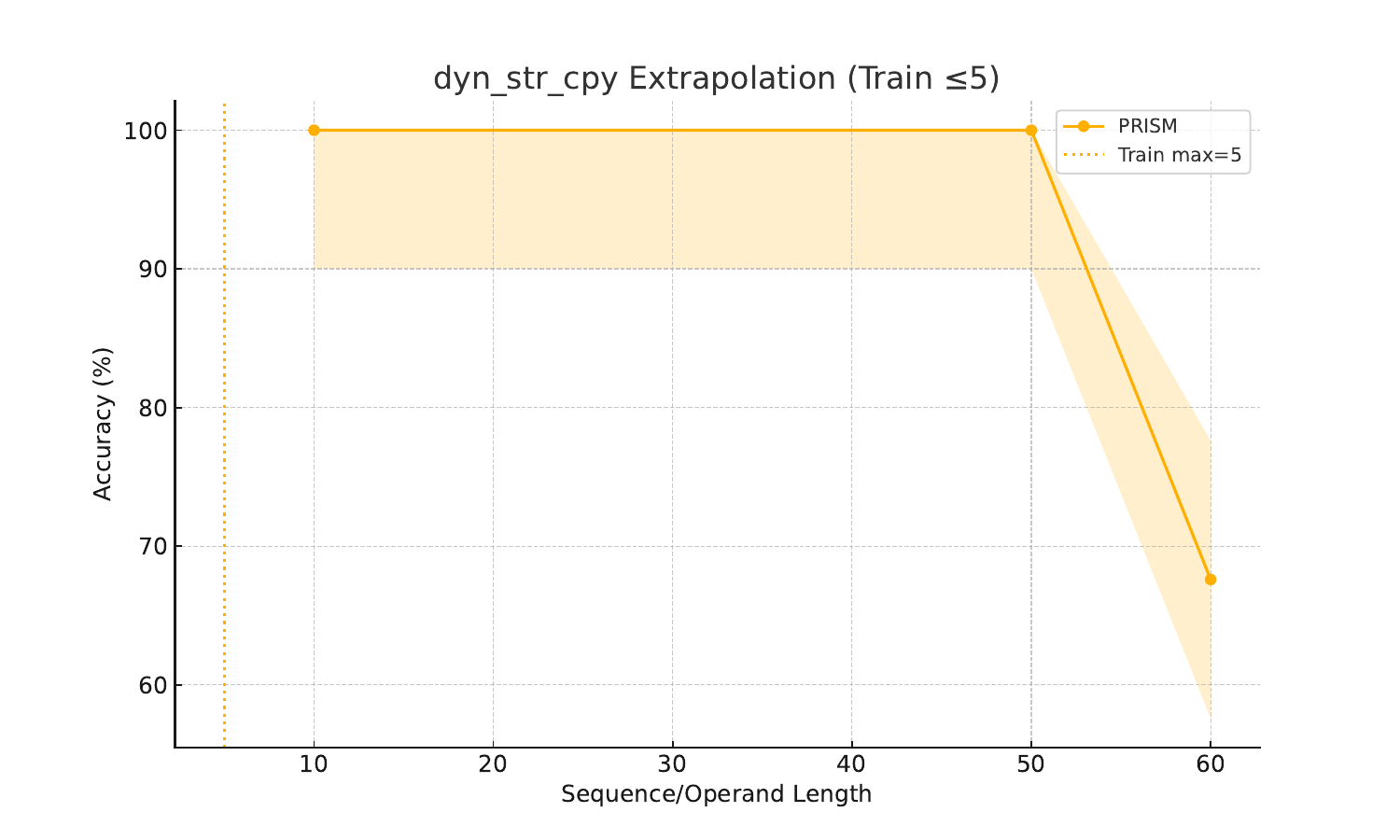}
  \caption{dyn\_str\_cpy extrapolation when trained up to length 5 (PRISM only).}
  \label{fig:dynstrcpy5}
\end{figure}

\begin{figure}[H]
  \centering
  \includegraphics[width=0.8\linewidth]{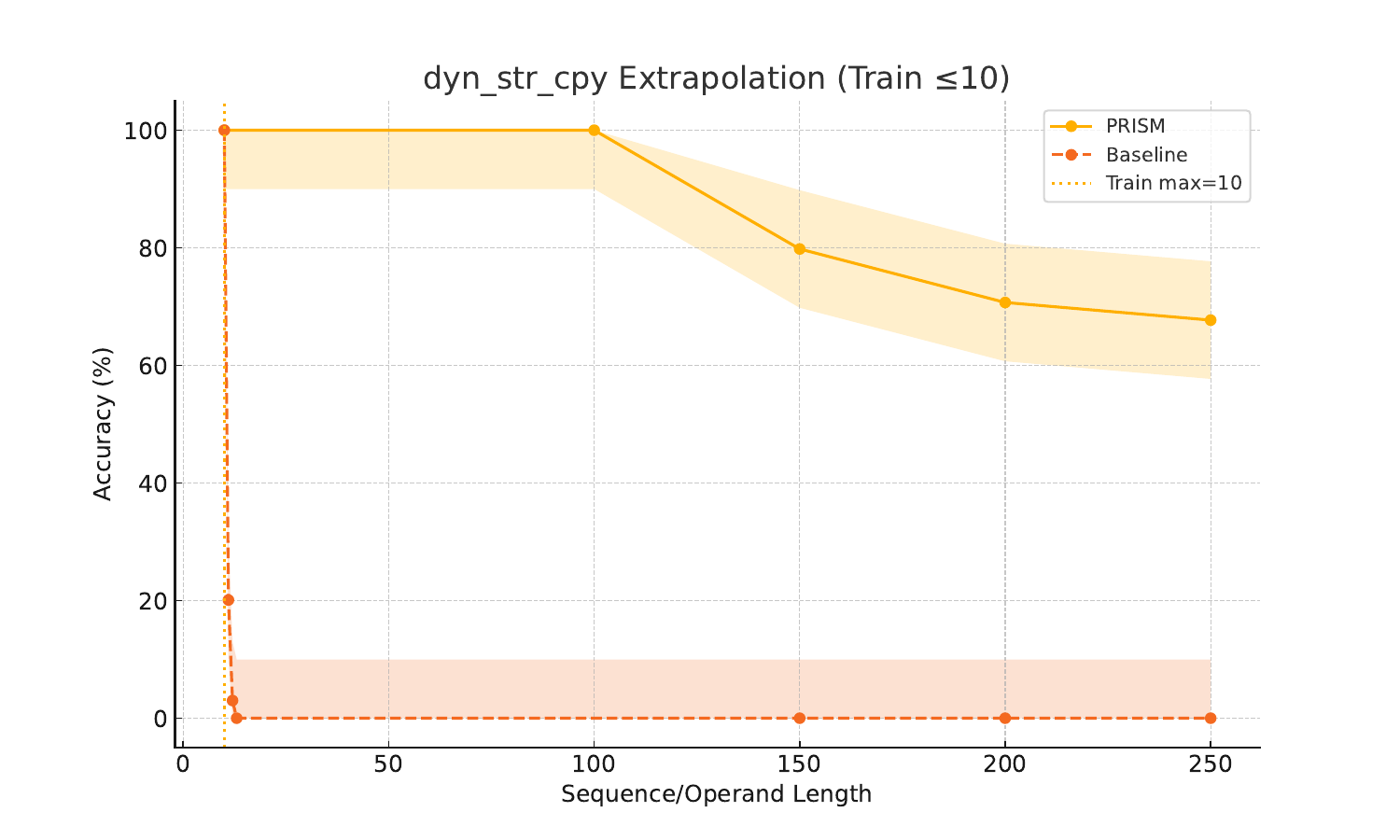}
  \caption{dyn\_str\_cpy extrapolation when trained up to length 10. PRISM vs.\ Baseline. Dashed line marks the training max length ($10$).}
  \label{fig:dynstrcpy10}
\end{figure}

\begin{figure}[H]
  \centering
  \includegraphics[width=0.8\linewidth]{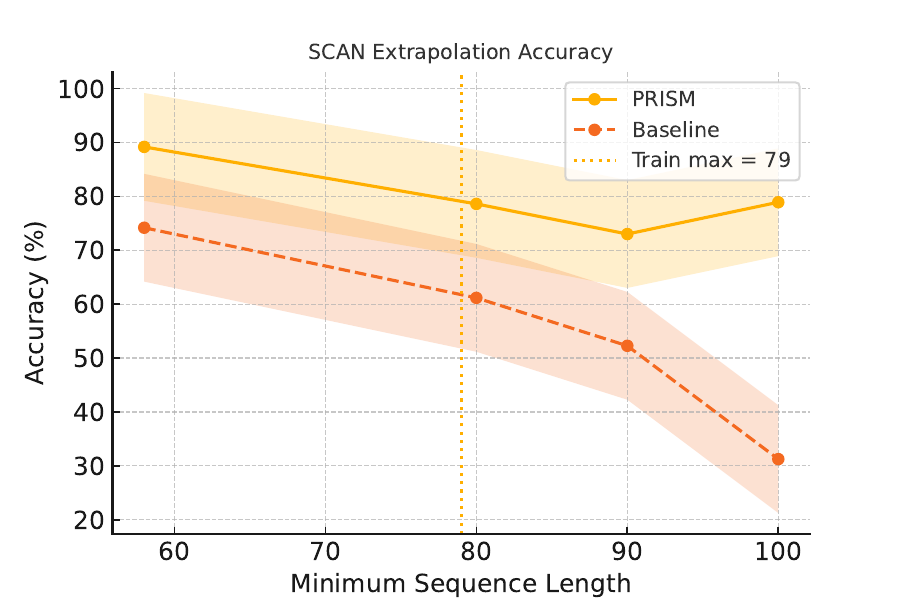}
  \caption{SCAN-CoT extrapolation accuracy for test dataset. Refer to Appendix~\ref{sec:scan-len-dataset-distr} for length distributions of train and test sets.}
  \label{fig:scan}
\end{figure}

\begin{figure}[H]
  \centering
  \includegraphics[width=0.8\linewidth]{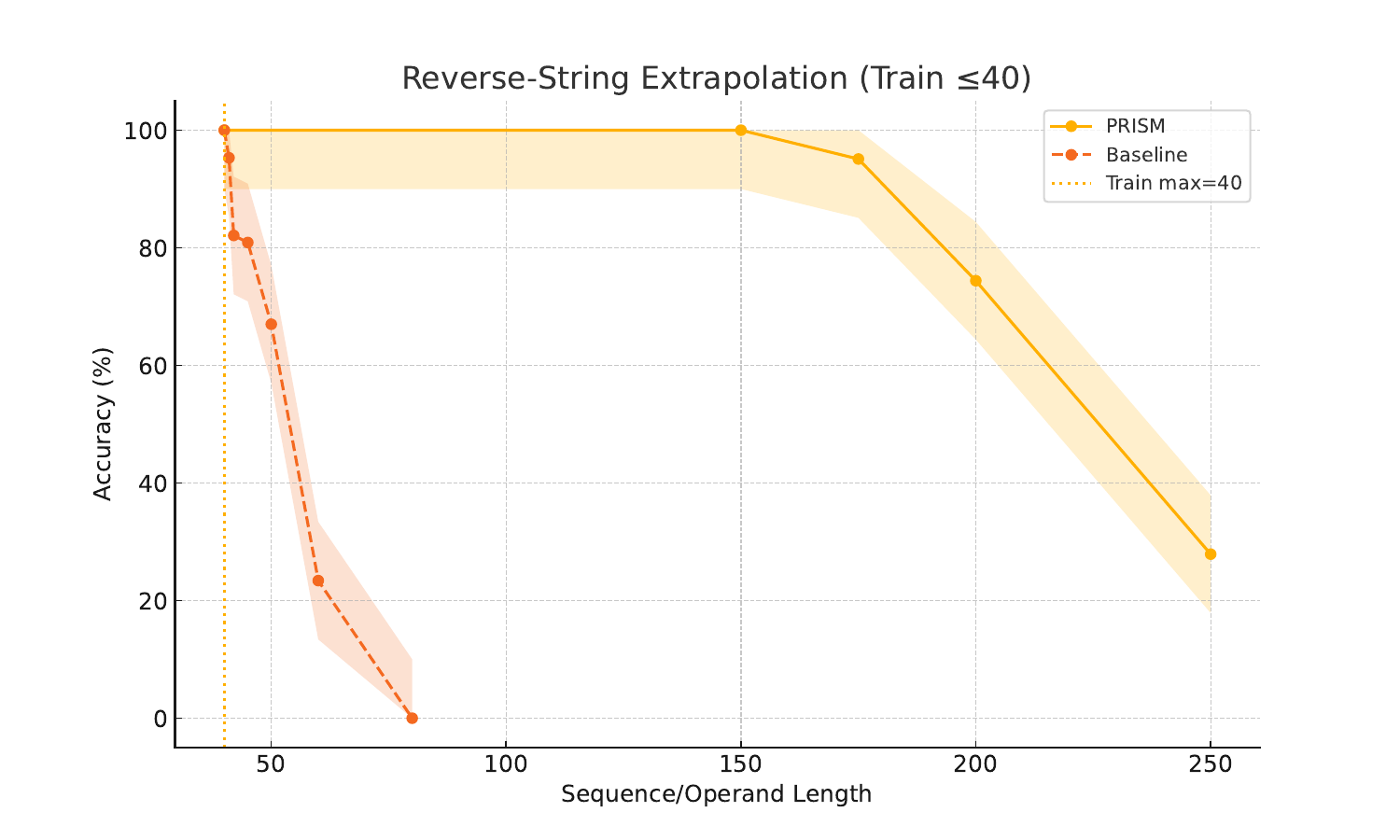}
  \caption{Reverse-String extrapolation for max‐train length $40$: PRISM vs.\ Baseline.}
  \label{fig:reverse40}
\end{figure}

\begin{figure}[H]
  \centering
  \includegraphics[width=0.8\linewidth]{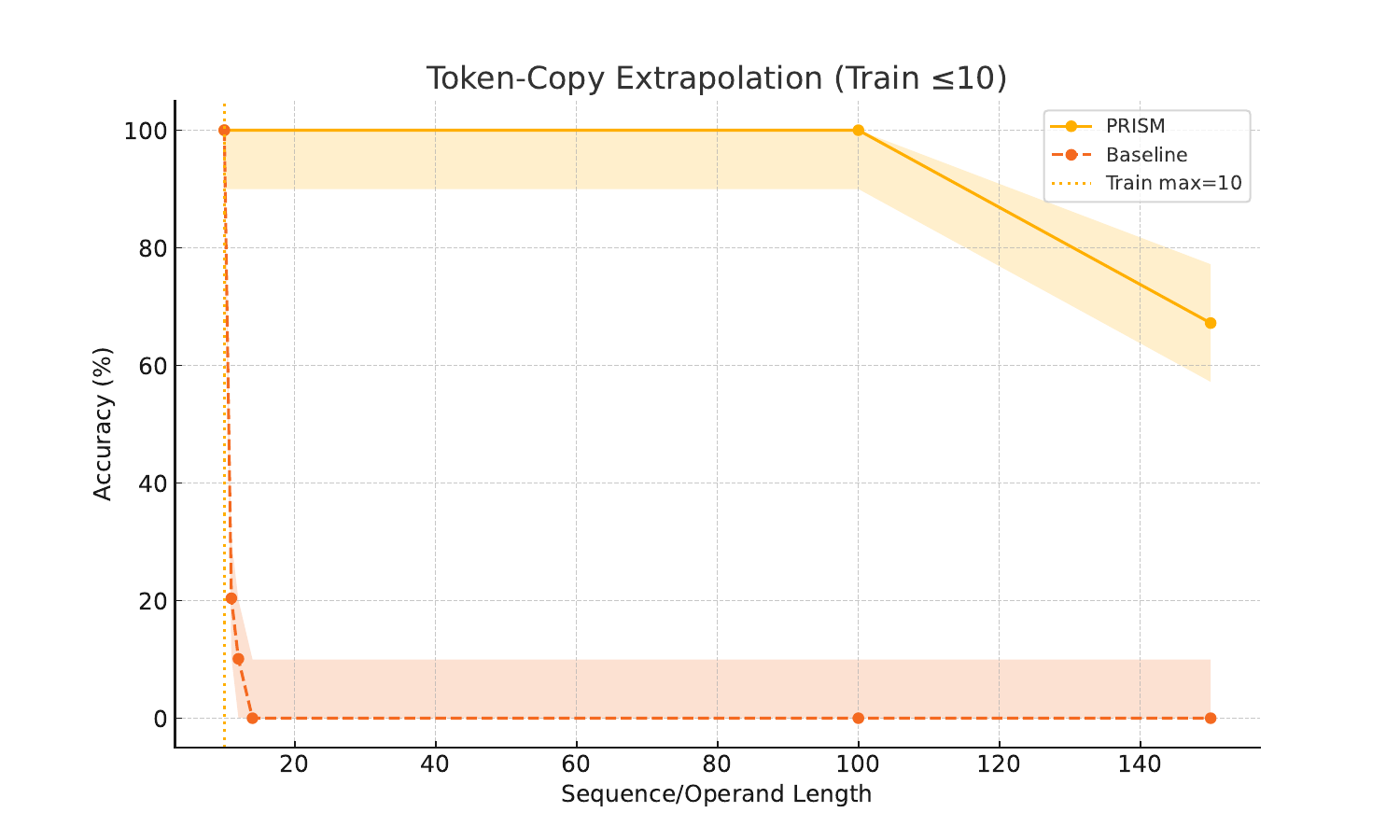}
  \caption{Token-Copy extrapolation when trained up to length 10: PRISM vs.\ Baseline.}
  \label{fig:tokencopy10}
\end{figure}

\begin{figure}[H]
  \centering
  \includegraphics[width=0.8\linewidth]{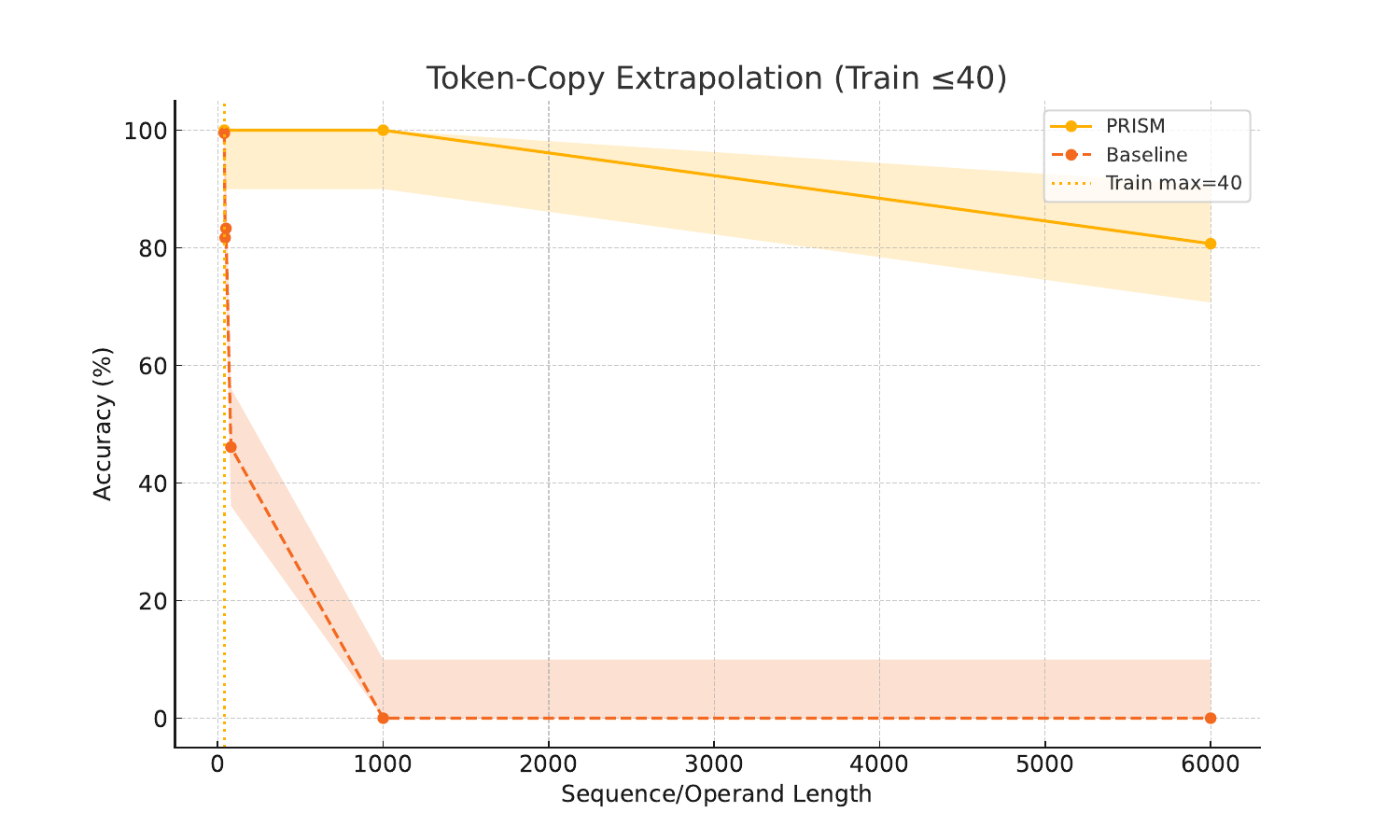}
  \caption{Token-Copy extrapolation when trained up to length 40: PRISM vs.\ Baseline.}
  \label{fig:tokencopy40}
\end{figure}

\begin{figure}[H]
  \centering
  \includegraphics[width=0.8\linewidth]{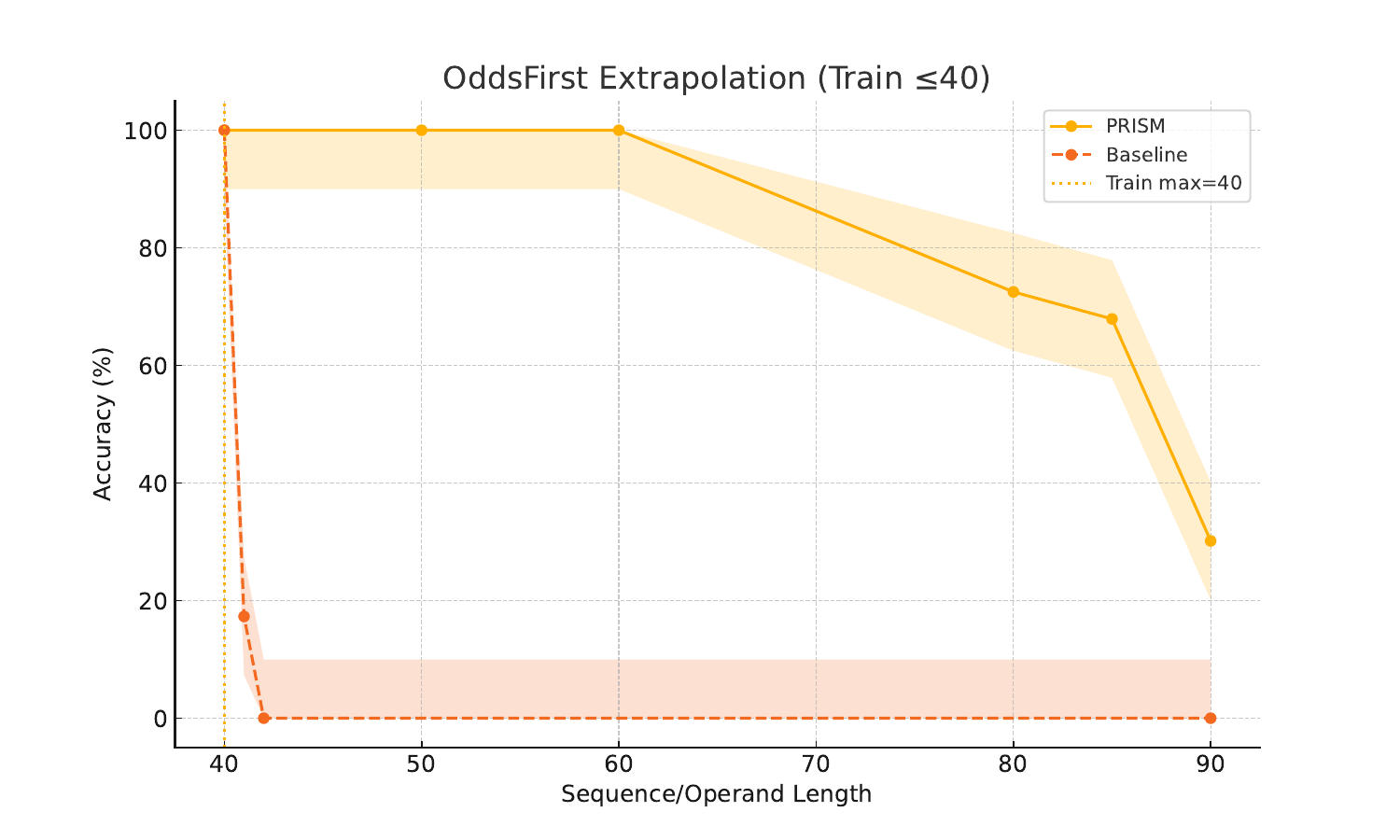}
  \caption{Odds First extrapolation accuracy for max‐train length $40$: PRISM vs.\ Baseline.}
  \label{fig:oddsfirst}
\end{figure}

\begin{figure}[H]
  \centering
  \includegraphics[width=0.8\linewidth]{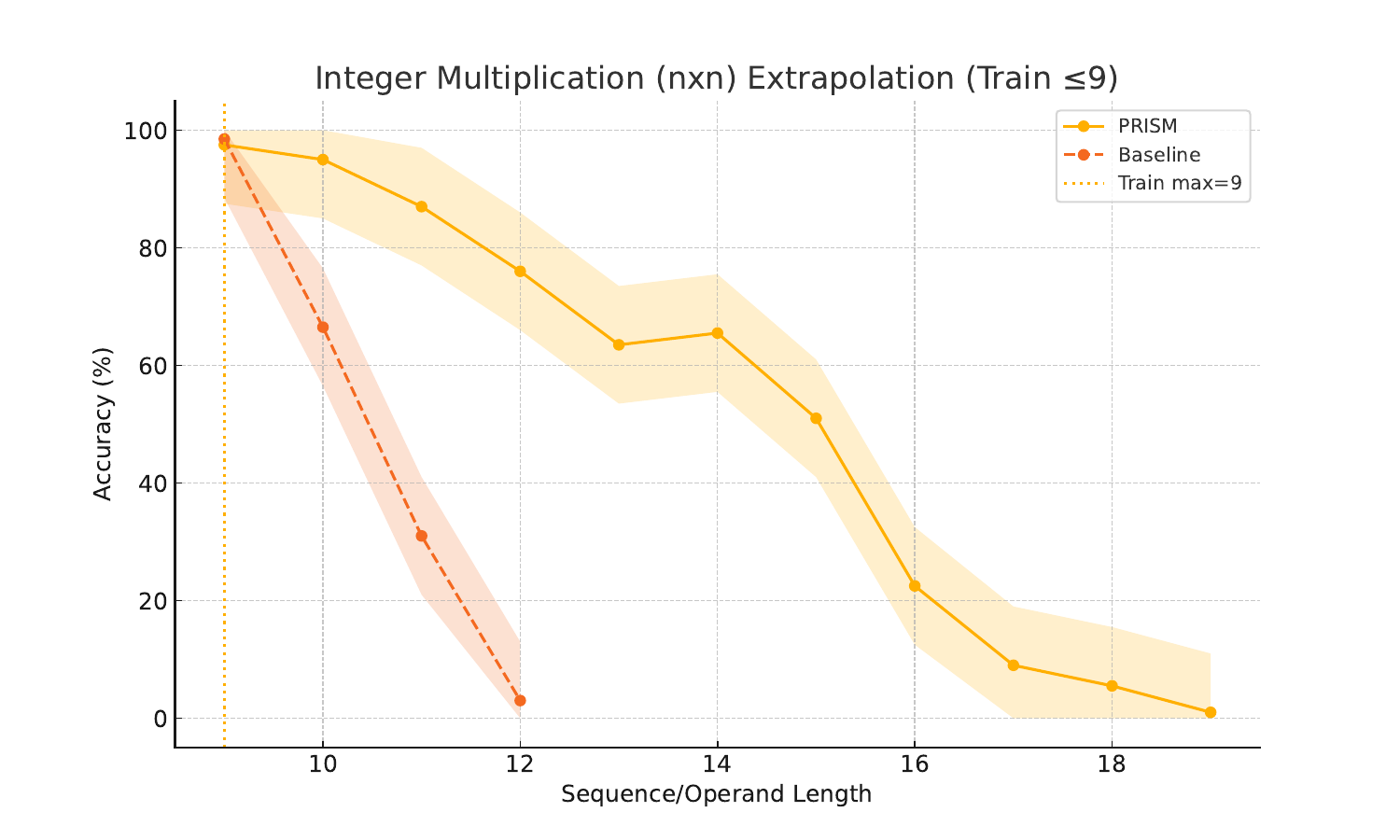}
  \caption{Integer Multiplication CoT (nxn) extrapolation for oper​and length up to 9: PRISM vs.\ Baseline. Dashed line marks the training max length ($9$).}
  \label{fig:multiplication}
\end{figure}

\begin{figure}[H]
  \centering
  \includegraphics[width=0.8\linewidth]{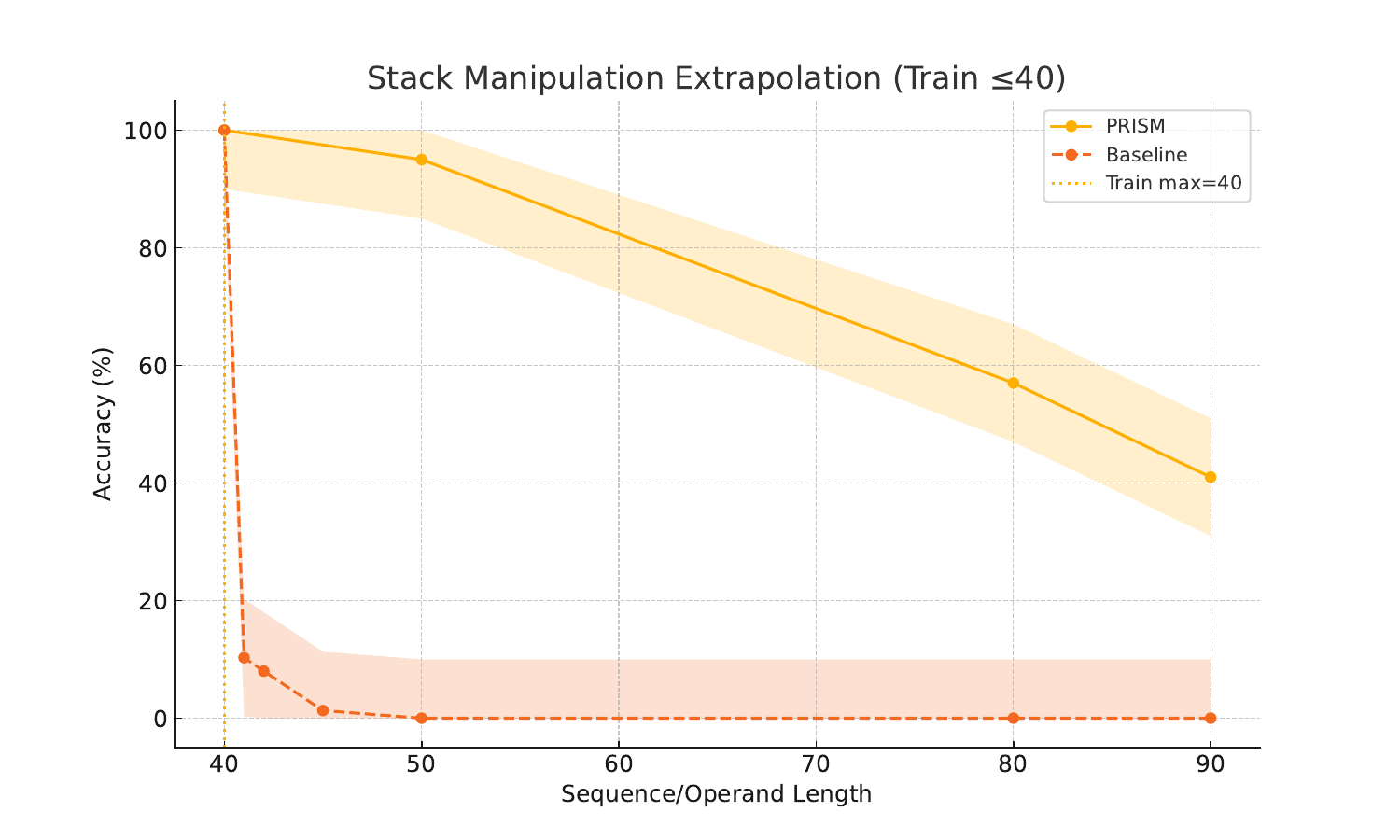}
  \caption{Stack Manipulation extrapolation for max‐train length $40$: PRISM vs.\ Baseline.}
  \label{fig:stackmanip}
\end{figure}

All models are evaluated every $1{,}000$ training steps (after crossing a loss threshold of $0.1$) on a held-out test set of $1{,}000$ examples.  For each evaluation point we record the accuracy, and in our final plots report the average of the \emph{top} three accuracies attained across training.  In Figures~\ref{fig:addition}–\ref{fig:stackmanip}, individual sample points (at each 1k step) are marked explicitly, and the dashed bands indicate a $\pm10\%$ margin around the line.

\subsection{Model Architectures}

\paragraph{Tokenization}
\noindent All models use per-digit tokenization and a vocab size of 64 across all tasks. On all tasks, the full vocab is not used but is kept constant.

\paragraph{Baseline Transformer.}
As a reference, our baseline is a standard 5-layer Transformer (Vaswani \emph{et al.} 2017) with sinusoidal absolute position encodings, 8 heads per layer, weight-tied unembeddings with the input vocab embeddings, and an embedding dimension of 512. We also use \textit{randomly shifted APE position encodings}~\cite{mcleish2024abacus} to allow the model to make use of all position embeddings.

\paragraph{Choosing a Representative Baseline for this Working Paper.}
Recent studies have consistently shown that a small number of positional-encoding schemes capture the bulk of length-extrapolation behavior in compact transformer models.  In particular,~\cite{ruoss2023random} demonstrated that randomized positional encodings (RPE) match or exceed the extrapolation accuracy of absolute positional encodings (APE) on a variety of synthetic sequence tasks.  McLeish \emph{et al.}~\cite{mcleish2024abacus} further showed that “Abacus” embeddings enable arithmetic transformers to generalize far beyond their training lengths with minimal modifications.  Kazemnejad \emph{et al.}~\cite{kazemnejad2023impact} conducted a broad comparison across multiple encoding families and found that, once hyperparameters are tuned, most alternatives perform on par with APE.  Finally, Saxton \emph{et al.}~\cite{saxton2019} analyzed arithmetic reasoning in small transformers and observed that augmenting APE with lightweight task-specific tweaks yields nearly identical length-extrapolation trends.  Taken together, these convergent results justify our focus on APE as a single representative baseline; we note, however, that an expanded empirical comparison—including RPE, Abacus, and other leading schemes—is underway and will appear in an upcoming revision.

\paragraph{PRISM + Transformer.}
We augment the vanilla transformer baseline with our PRISM layer(s) as follows:
\begin{itemize}
  \item \textbf{All tasks except Odds First and Stack Manipulation:} a single PRISM layer is inserted \emph{before} the first Transformer layer.
  \item \textbf{Odds First and Stack Manipulation:} two PRISM layers are used, one before layer,1 and one before layer,4.
  \item Remove the absolute position encoding by decoupling the APE from the content embedding residual stream, which partially draws inspiration from the disentanglement from~\cite{he2021deberta}, except that we formally take the superposition of \textit{all} residual stream~\cite{anthropic2023privileged} position encoding vectors via concatenation and linear combination.
  \item Unless otherwise noted, the total number of \textit{Histogram and corresponding superposition-PE Cursors} is set to 140, with $c$=4 cursors and {key, query} pairs per head. Refer to Section~\ref{sec:hierarchical-pe} for details.
\end{itemize}
In all cases the sinusoidal superposition embedding dimension is 340, the token–vocabulary embedding dimension is 192, and the Transformer has 5 layers.

\subsection{Regularization}
For all tasks (unless otherwise noted), we enforce orthogonality on the GRU hidden‐to‐hidden matrix $W_{hh}$ via one of:
\begin{itemize}
  \item PyTorch parametrization: 
    \texttt{from torch.nn.utils.parametrizations import orthogonal}, or
  \item an explicit Frobenius‐norm regularizer of weight $10^{-3}$ on $W_{hh}$.
\end{itemize}
\footnotetext{* In subsequent ablations we found that normalizing the Frobenius norm of $W_{hh}$ to be a fixed multiple of a small learnable constant further improved extrapolation on Odds First, SCAN, and (in preliminary experiments) most other tasks.}

\subsection{Training details}
The PRISM architecture is trained with AdamW ($\beta_1{=}0.9,\beta_2{=}0.98$),
for a maximum of 150k steps to a peak LR of $9{\times}10^{-5}$, batch
size~$100$, sequence length capped at $2048$ tokens (padding where
necessary). All baselines are trained for 150k steps with a batch size of 100, unless otherwise noted.
Across all tasks, we warm up using the minimum sequence lengths (up to
length~5) for 5 k steps, then train using a maximum sequence length of 10
for the next 5 k steps.
\textbf{On Stack Manipulation, Reverse, Odds First, and CoT Addition we increase the maximum sequence length in steps of 10, allocating 10 k
steps to each length bucket.}
For the $\alpha$ coefficients (§ \ref{sec:alpha-coeff}) of our relative
position cursors we use a separate AdamW group
($\beta_1{=}0.8,\beta_2{=}0.92$, $\text{lr}{=}0.03$).

\section{Results}\label{sec:results}
All charts follow the Experimental setup (Section~\ref{sec:exp-setup}) and use the average‐of‐top‐3 sampling method for reporting accuracies. Unless otherwise stated, dashed vertical lines represent the maximum training distribution for each graph.

\section{Discussion}

\noindent We start from the empirically validated view~\cite{kazemnejad2023impact, Zhou2024algorithms,DziriFaithAndFate2024,Deletang2023Chomsky,cho2024coupled,golowich2025sparsity,ruoss2023random,mcleish2024abacus} that a wide range of simple and important skills, requiring \textit{algorithmic} or \textit{long-context} reasoning traces, may be set back by the limitations of existing neural network architectures. These tasks are easy for humans to learn via generalizing from observations of correct examples, so that they (a) can execute the same tasks except where the time horizon extends beyond that observed during training, and (b) execute tasks via \textit{compositionality} - chaining together very basic rules as the building blocks of a more complex task. In this paper, we have focused on both forms of generalization for neural networks via training them to learn from self-supervised data.
\newline
\newline
\noindent 
\noindent The limited \textit{length generalization} on simple tasks that are otherwise the same as learned during training, except extended beyond the training horizon - which are limitations demonstrated by transformer, recurrent, and other sequence-to-sequence architectures - motivates this work. \\
\newline
\noindent Before a model can execute complex, practical tasks beyond the training horizon, and do so \textit{reliably} and \textit{autonomously}, the model must first have the ability to execute simpler tasks or algorithms. Practical, complex tasks include programming, software design, software modification, planning, mathematical proofs. We reiterate that what they all have in common is compositional reasoning, i.e. \textbf{applying the same basic rules and subroutines} that the model has learned in order to break down any complicated task~\cite{DziriFaithAndFate2024} into simpler steps~\cite{lake2018}. Fundamentally, these simpler steps require \textbf{length generalization} of the language model if the time horizon is increased beyond the training distribution. \\

\noindent Our experiments confirm that the proposed relative-position stream
extrapolates far beyond the training length on tasks that defeat
standard Transformer baselines—e.g.\ 20-digit chain-of-thought
addition after training on only 10 digits, or copying/reversing
100-token strings after seeing 10-token examples.  These findings
support our long-term goal of \emph{sequence-length generalization}
for language-model reasoning. \\

% \newline

\noindent Inspired by the view that scaling up the reliability of LLMs requires learning from data (i.e. self-supervision)~\cite{Sutton2019BitterLesson} and from the agent's own experiences - to the best of our knowledge - we are first to demonstrate length generalization on a variety of simple algorithmic reasoning tasks, like addition via chain-of-thought, using a purely \textbf{\textit{self-supervised} approach}, meaning that our model can, in principle, learn the same tasks from a text corpus without the use of hand-crafted labels or other forms of supervised learning. Notably, other architectures trained using self-supervision have previously been noted to fail to extrapolate altogether, even via chain-of-thought prompting or other scratchpads - necessitating approaches like providing \textit{index hints} during training and inference.  \\

% \linebreak

\noindent Demonstrating the understanding of simple algorithms by tracing all of their \textit{required steps}, such as via our Chain-of-Thought Addition task, establishes confidence that \textbf{arbitrary, more complex algorithms} may be generalized beyond the training horizon. For example, algorithms are, in general, similarly unbounded in their length and memory requirements, just as an algorithm that computes the sum of two numbers can be arbitrarily complex with respect to the amount of computation steps required. Thus, when understanding \textit{any} algorithm it has never seen before, an AI system needs to perform a sequence of operations described by a set of instructions. Thus, we view length extrapolation on the simple algorithms explored in this paper as prerequisites to extrapolation on not only more complicated forms of \textit{algorithmic reasoning}, like graph traversal, but also tasks including software engineering \& algorithm design (computer science research), systematically resolving interdependencies when building a software package (dev-ops), and in general, any task that can be broken down into simpler steps in order to describe a \textbf{correct reasoning trace} or \textbf{plan of how to execute the task}. \\

\noindent Because of the above and PRISM's general-purpose architecture, which can be added to any transformer layer, we view our achievements as a stepping stone to models reliably transcending the training horizon on practical tasks.

\section{Future Work}

\noindent Aside from long-context tasks involving \textit{sequential data} - a model should also be able to perform reasoning involving other modalities, such as graphs (e.g. a network of roads, tree-structured data), diagrams (e.g. UML, concept maps), spatial reasoning, and visual flow charts. While tasks involving these modalities similarly involve the step-by-step execution of a plan, the observations are not sequential in general. \\

\noindent However, due to the understanding of the spatial structure of the data that our neural architecture has demonstrated, we leave the exploration of tasks that involve the understanding of these general modalities for future work.

\appendix
\section{Briefly Justifying Architectural Design Decisions}

\subsection{Why not ``index hints’’ or similar, programmed solutions?}
Prior methods such as \emph{Abacus Embeddings}, \textit{Inductive Scratchpads}, and related works attach fixed, one-hot embeddings at special positions (digit boundaries, start/end tokens, etc.), effectively implementing a \textit{predetermined} counting-based \textit{reset} scheme in a hard-coded, monotonically increasing fashion.
Because the landmarks are \emph{not learned}, these schemes do not
induce a distance metric nor a similarity function: attention fires only
when the two tokens carry the \emph{exact same} hint, ignoring intermediate
positions and weakening generalization once the presence of hard-coded features, sequence, or the
copy-window length change.  Consequently they
isolate the “correct’’ token by construction but do not learn a genuine
\emph{relative position encoding} that satisfies the semantics of a distance metric.  Such hard-coded features boost
in-distribution accuracy on short, template-like tasks, yet they break
down on compositional or chain-of-thought settings and cannot
generalize once the latent layout differs from the hand-coded pattern.
In contrast, our model learns the metric \(\,d(\cdot)\,\) and its
similarity map directly from data through the stochastic histogram
update, enabling smooth extrapolation without manual anchors.\\

\noindent Our approach achieves this by \emph{learning}
both the metric \(d\) and the coefficients of the sinusoidal embedding inner product when computing the unnormalized attention matrix, allowing smooth,
data-driven extrapolation to sequences far longer than those observed
during training.

\vspace{0.5em}
\noindent

% \subsection{Architectural Design Choices}

% \paragraph{Stochastic-matrix view.}
% The histogram update is mathematically equivalent to marginalizing a
% $T$-step categorical branching process with $O(P)$ rather
% than $O(P^{2})$ time per step.  Concretely, we factored the transition matrix $P\times P$ into three rank-1 blocks
% (\emph{reset}, \emph{increment}, \emph{decrement/keep}),
% reducing the complexity from $O(P^{2})$ to $O(P)$ while preserving exact
% marginals.  

\subsection{Sum and Sharpen using the ${\gamma}$ parameter.}\label{app:power-sharpen}

To counteract the exponential damping of confident local gates
(e.g.\ $0.99^{50}\approx0.60$ after 50 steps), we apply a \emph{power-sharpening} nonlinearity described in \ref{sec:gamma-sharpen}:
\[
\bm{h}_{t+1}^{(\mathrm{sharp},\,c)}[i]
\;=\;
\frac{\bigl([\bm{h}_{t+1}^{(c)}]_i + \varepsilon\bigr)^{\gamma_c}}
     {\sum_j \bigl([\bm{h}_{t+1}^{(c)}]_j + \varepsilon\bigr)^{\gamma_c}},
\]
where each cursor histogram \(c\) has its own learnable exponent \(\gamma_c\),
initialized to \(2.0\) at model creation.

\subsubsection{Markov-kernel perspective on the “cohesion” effect of ${\gamma}$.}
Equivalently, each sharpen-then-renormalize step can be seen as applying a
reweighted one-step transition matrix with temperature
\(\tau_c = 1/\gamma_c < 1\).  Over multiple timesteps this amounts to
repeated exponentiation of the same stationary kernel,
sharpening high-probability transitions and thereby promoting “cohesion”
rather than diffusion.

\begin{proof}
We prove by induction on \(n\) that for a fixed sharpening exponent \(\gamma_c\),
\[
p_n(i)
=\frac{p_0(i)^{\gamma_c^n}}
      {\displaystyle\sum_{j}p_0(j)^{\gamma_c^n}},
\]
where each update is
\[
p_{t+1}(i)
=\frac{p_t(i)^{\gamma_c}}{\displaystyle\sum_{j}p_t(j)^{\gamma_c}}.
\]

\medskip

\textbf{Base case (\(n=0\)).}
Since \(\gamma_c^0=1\),
\[
\frac{p_0(i)^{\gamma_c^0}}
     {\sum_{j}p_0(j)^{\gamma_c^0}}
=
\frac{p_0(i)}{\sum_{j}p_0(j)}
= p_0(i),
\]
because \(p_0\) is already normalized.

\medskip

\textbf{Inductive step.}
Assume the claim holds for \(n=k\), and let
\[
S_k = \sum_{\ell}p_0(\ell)^{\gamma_c^k},
\qquad
p_k(i)=\frac{p_0(i)^{\gamma_c^k}}{S_k}.
\]
Then
\[
p_{k+1}(i)
= \frac{p_k(i)^{\gamma_c}}{\sum_{j}p_k(j)^{\gamma_c}}
= \frac{\bigl(\tfrac{p_0(i)^{\gamma_c^k}}{S_k}\bigr)^{\gamma_c}}
       {\displaystyle\sum_{j}\bigl(\tfrac{p_0(j)^{\gamma_c^k}}{S_k}\bigr)^{\gamma_c}}
= \frac{p_0(i)^{\gamma_c^{k+1}}/S_k^{\gamma_c}}{\sum_{j}p_0(j)^{\gamma_c^{k+1}}/S_k^{\gamma_c}}
= \frac{p_0(i)^{\gamma_c^{k+1}}}{\sum_{j}p_0(j)^{\gamma_c^{k+1}}},
\]
completing the induction.

\medskip

Thus after \(n\) repeated sharpenings, the original distribution’s exponents have accumulated to \(\gamma_c^n\), rigorously capturing the “cohesion” effect.
\end{proof}

\subsection{Separate "Position Stream" and Linear Combinations of \textit{Sinusoidal Position Encoding}}\label{convex-combinations-instead-of-probabilistic-mixture}  
Crucially, instead of a probabilistic mixture distribution, the "probability distribution" represented by our histogram (\ref{sec:histogram-update-rule-highlevel-descr}) $h_{t}$ is more correctly interpreted as the coefficients of a linear (convex) combination of sinusoidal PEs at each time-step. This is in the same way that the self-attention mechanism of~\cite{vaswani2017} takes the convex combination of value vectors. Therefore, in practice, we represent the "posterior distribution"
over relative positions $\mathbf{h_{t}}$ as a superposition
of \textbf{sinusoidal positional encoding vectors}.\\

\noindent In other words, for a vector dimension $d_{model}$ and position index $k\in\{0,\dots,P{-}1\}$, we use the
``Transformer-classic’’ encoding introduced from~\cite{vaswani2017}.
Averaging scalars would erase phase information that these
encodings need for dot–product attention.
Indeed, for any frequency~$\omega_{m}$ in the basis,
the similarity between two positions satisfies
\[
  \bigl\langle f_{\omega_{m}}(k),\,f_{\omega_{m}}(\ell)\bigr\rangle
  \;=\;
  \cos\!\bigl(\omega_{m}(k-\ell)\bigr),
\]
which \emph{decays smoothly} as the relative distance
$d(k,\ell)=|k-\ell|$ grows, while exhibiting a sharp, Gaussian-like
peak at $k=\ell$ when the sum over multiple frequencies
is taken.  This graceful fall‐off provides fine spatial
resolution locally and coarse invariance at larger offsets—crucial
for robust long-context reasoning.

\paragraph{Superposition and Relative Position Encoding}

Let $f:\{0,\dots,P{-}1\}\!\to\!\mathbb{R}^{d}$ be the sinusoidal position encoding and let $\bm{h}_{t}[k]$ denote the \emph{probability} that the latent cursor is at position $k$ after step~$t$.
We represent the uncertainty by a \emph{superposition} of
encoding vectors,
\begin{equation}
\bm{e}_{t} \;=\;
\sum_{k=0}^{P-1} \bm{h}_{t}[k]\;f(k)
\;\;\in\; \mathbb{R}^{d}.
\label{eq:superposition}
\end{equation}

\paragraph{Why not average scalars first?}
Suppose one tried to turn the histogram into a single \emph{expected position}
$\bar{k}_{t}\!=\!\sum_{k} k\,\bm{h}_{t}[k]$ and then feed
$f(\bar{k}_{t})$ downstream.
Because $f(\cdot)$ is \emph{non-linear}
(component-wise sinusoid or MLP),
\[
f\!\left(\,\bar{k}_{t}\,\right)
\;\neq\;
\sum_{k} \bm{h}_{t}[k]\,f(k)
\quad
\text{in general}.
\]
The mismatch is severe in high-frequency dimensions.
For the sinusoidal encoder $f_{2m}(k)=\sin(\omega_{m}k)$,
$f_{2m+1}(k)=\cos(\omega_{m}k)$,
we have
\begin{align*}
f_{2m}\!\left(\bar{k}_{t}\right)
&=\sin\!\bigl(\omega_{m}\bar{k}_{t}\bigr)
\;\;\;\text{vs.}\;\;
\sum_{k} \bm{h}_{t}[k]\sin(\omega_{m}k)
     \;=\;\Im\!\Bigl\{\mathcal{F}_{\omega_{m}}\bigl[\bm{h}_{t}\bigr]\Bigr\},\\
f_{2m+1}\!\left(\bar{k}_{t}\right)
&=\cos\!\bigl(\omega_{m}\bar{k}_{t}\bigr)
\;\;\;\text{vs.}\;\;
\sum_{k} \bm{h}_{t}[k]\cos(\omega_{m}k)
     \;=\;\Re\!\Bigl\{\mathcal{F}_{\omega_{m}}\bigl[\bm{h}_{t}\bigr]\Bigr\},
\end{align*}
where $\mathcal{F}_{\omega_{m}}$ is the discrete Fourier coefficient at
frequency~$\omega_{m}$.\\

\noindent The Fourier transform of \emph{the whole histogram} appears on the
right-hand side, whereas the left collapses everything to one phase.
Consequently, averaging positions destroys interference patterns that are
crucial for attention to disambiguate periodic cues (e.g.\ copy-tasks,
music, code indentation).

\paragraph{Summary}
Equation~(\ref{eq:superposition}) ensures that
every plausible position contributes its \emph{full} encoding,
weighted by its probability.
The model can therefore
\emph{adaptively focus} on sharp or diffuse spatial hypotheses simply by
modulating $\bm{h}_{t}$, \emph{without} any change to the downstream
attention mechanism.
This maintains expressiveness while retaining the
$\mathcal{O}(P)$ update cost established in
Section~\ref{sec:computational-complexity}.

\subsection{Reset Transition - Additional Justifications}
\label{sec:dynamic-origin-other-interpretations}

\noindent The reset gate (\ref{eq:pmf_with_reset}) \textit{precedes} increment/decrement, matching human procedural
        reasoning (``Assign a distance function relative to the left-most digit’’) and
        avoids mutually-exclusive gate assumptions that
        impede training when both $\mathrm{inc}$ and
        $\mathrm{dec}$ are plausible.\\

\paragraph{Higher-dimensional interpretation.}
\noindent Another interpretation is that in 2 and higher dimensions, a reset at some (spatial) position \textbf{t} corresponds to centering the \emph{coordinate frame} (and setting the distance function ${d = 0}$) at a landmark at location \textbf{t}, allowing the relative distance from the landmark to be correctly computed (similarly via a learned distance field) by path-integrating its gradient.\\

\noindent In other words for $\mathbb{R}^{n}$ ($n\!>\!1$), a reset at spatial landmark
$\bm{t}\in\mathbb{R}^{n}$ sets the scalar potential
$\phi(\bm{x})$ to zero at $\bm{x}=\bm{t}$.
\noindent The reset gate therefore \emph{dynamically centers the coordinate
frame}, ensuring that the learned distance function behaves
consistently—preventing scenarios (e.g.\ repeated landmarks) where a naïve difference $r_{t'}-r_{t}$ would be ill-defined.

\vspace{0.5em}
\noindent
\subsection{Summary}
\begin{itemize}
    \item Collectively, these inductive biases generalize classic discrete state Markov models: the transition matrix is parameterized by learnable gate probabilities at each step. 
    \item We can view the model as a \emph{linear–chain conditional random field} (CRF) whose edge potentials are produced by the neural gates.  
    \item The forward messages are updated in Eq.~(\ref{sec:histogram-update-rule-highlevel-descr}) via a sparse matrix–vector multiply that costs only $\mathcal{O}(P)$, instead of the naive $\mathcal{O}(P^{2})$ required by a dense transition matrix.
    \item The model's state at each time/position $t$ is a categorical distribution over \textit{relative} positions $\mathbf{h_{t}}^{(c)}$, rather than a single scalar or one-hot. 
    \item The categorical distribution $\mathbf{h_{t}}^{(c)}$ at each timestep is a vector of coefficients for the \textit{convex combination} of position encoding vectors.
    \item In particular, this convex combination emits a position encoding vector $\mathbf{e_{t}}^{c}$ via multiplying the categorical distribution vector $\mathbf{h_{t}}^{(c)}$ with the standard APE\footnote{Importantly, despite the use of an "APE" table, all emitted PE vectors for our learned model represent \textit{relative position.}} matrix from~\cite{vaswani2017} at each timestep/position $t$. 
    \item This emitted (linearly combined) PE vector $\mathbf{e_{t}^{c}}$ is then assigned to a unique orthogonal subspace of the \textit{position stream}, where this orthogonal subspace is uniquely determined via \textit{concatenation} according to the cursor index $c$ of the histogram $\mathbf{h_{t}}^{(c)}$; this step is described in~\ref{eq:similarity}.
    \item The update resembles a generalized Bernoulli trial at each step, but with non-exclusive, learned gate activations.
    \item Although our only focus is extrapolation, we take inspiration from~\cite{nerf2020} for learning a signed distance field, which can then be integrated. Their work generalized well on \textit{interpolation} for view synthesis between sampling points where later work~\cite{tancik2020fourier} has shown that sinusoidal position encodings for the observation kernel have allowed for higher frequency features to be accurately learned for reconstruction.
\end{itemize}

\section{Interpretability Visualizations}
\label{app:interpretability}

\begin{figure}[H]
  
  \centering
  \includegraphics[width=\linewidth]{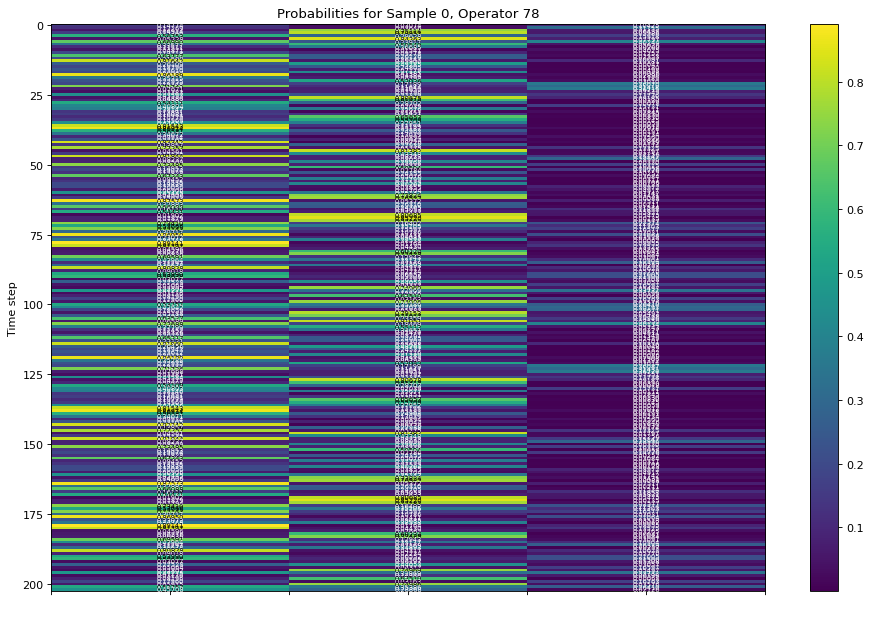}

  % ┌───────────────────────────────────────────────────────────────────┐
  % │  Use [ … ] to give a “short” caption without fragile bits.       │
  % │  Then, in the mandatory { … } (the “long” caption),              │
  % │  wrap \footnotemark in \protect so it doesn’t break the parsing. │
  % └───────────────────────────────────────────────────────────────────┘
  \caption[Gate probability distributions at initialization]{
    Gate probability distributions at initialization. Each PRISM cursor starts with uniform priors over the four actions—increment, decrement\protect\footnotemark, and reset—reflecting maximal initial uncertainty about position updates. Columns from left to right correspond to {increment, decrement, reset}, respectively, for the histogram probabilities at each position defined in~\ref{eq:pmf_with_reset}.
  }
  \label{fig:init-gate-probs}

\end{figure}

\begin{figure}[H]
  \centering
  \includegraphics[width=\linewidth]{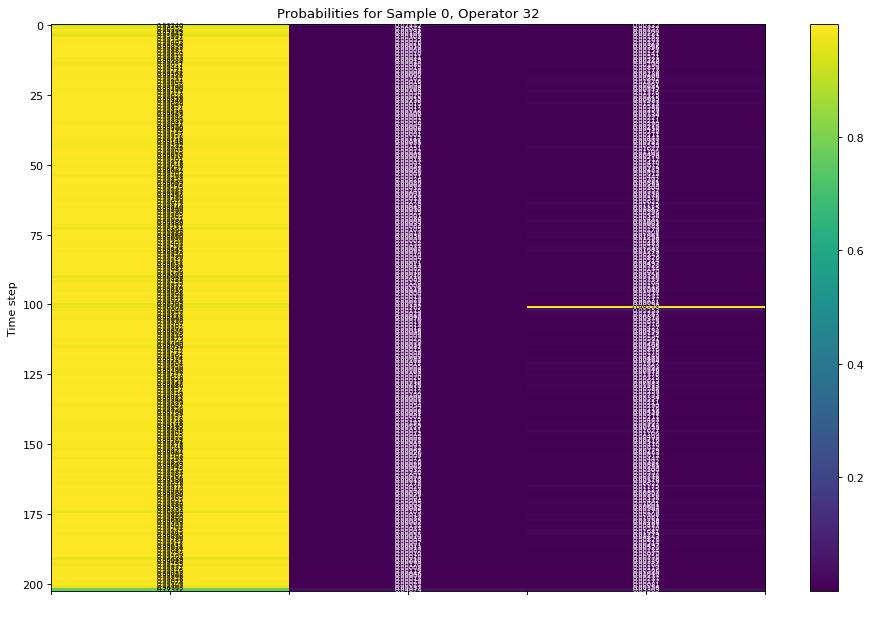}
  \caption{
  % \textbf{Figure A.2:} 
  Learned gate probability distributions after training on token-copy.  PRISM develops a strong bias toward \emph{increment} actions, with resets and decrements used sparingly, enabling smooth advancement of the internal cursor across long sequences.}
  \label{fig:trained-gate-probs}
\end{figure}

\begin{figure}[H]
  \centering
  \includegraphics[width=\linewidth]{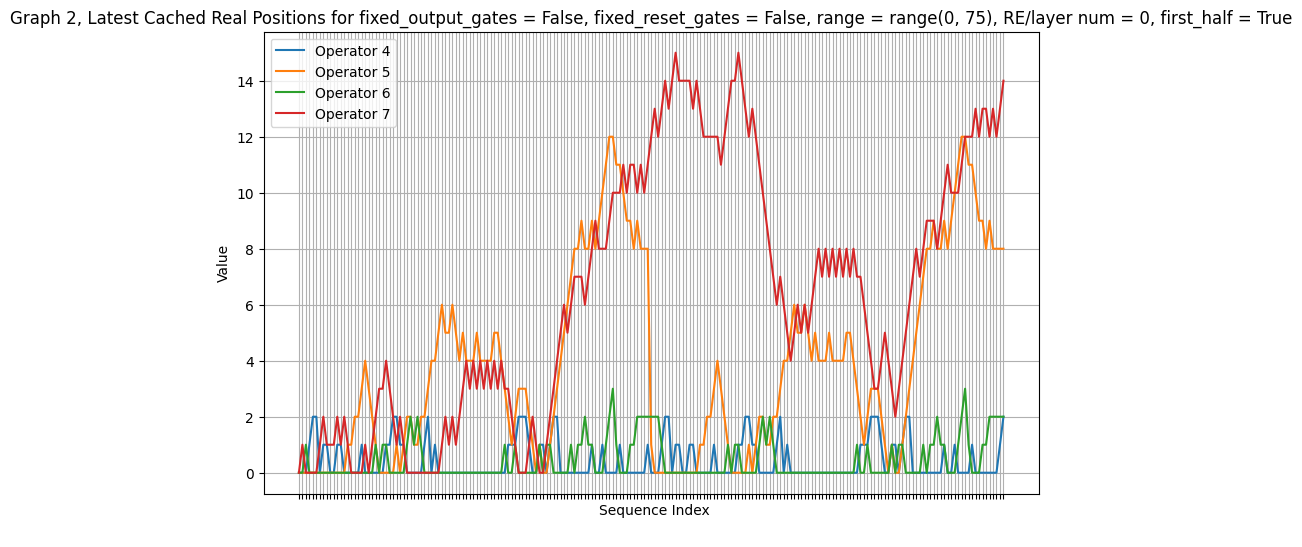}
  \caption{
  % \textbf{Figure A.3:}
  Token‐copy task, \emph{initialization}.  The top-1 mode of the relative‐position histogram is diffuse and nearly uniform prior to training, indicating that the model has no initial bias about which past position to copy from.}
  \label{fig:tokencopy-init}
\end{figure}

\begin{figure}[H]
  \centering
  \includegraphics[width=\linewidth]{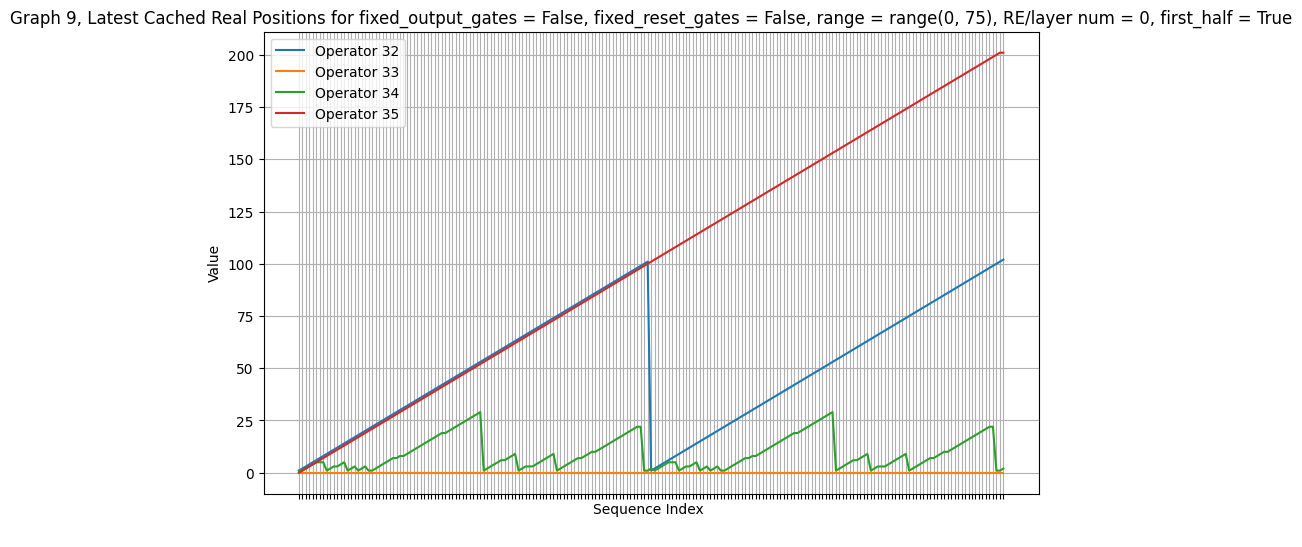}
  \caption{
  % \textbf{Figure A.4:}
  Token‐copy task, \emph{after training}.  The histogram sharpens dramatically, with the blue query cursor's mode exactly matching the true offset between each query and its corresponding key—demonstrating precise learned localization. Note that both query and key histograms are learned from scratch from random initialization, demonstrating an emergent coupling of the keys to \textit{uniquely align} with the queries.}
  \label{fig:tokencopy-trained}
\end{figure}

\begin{figure}[H]
  \centering
  \includegraphics[width=\linewidth]{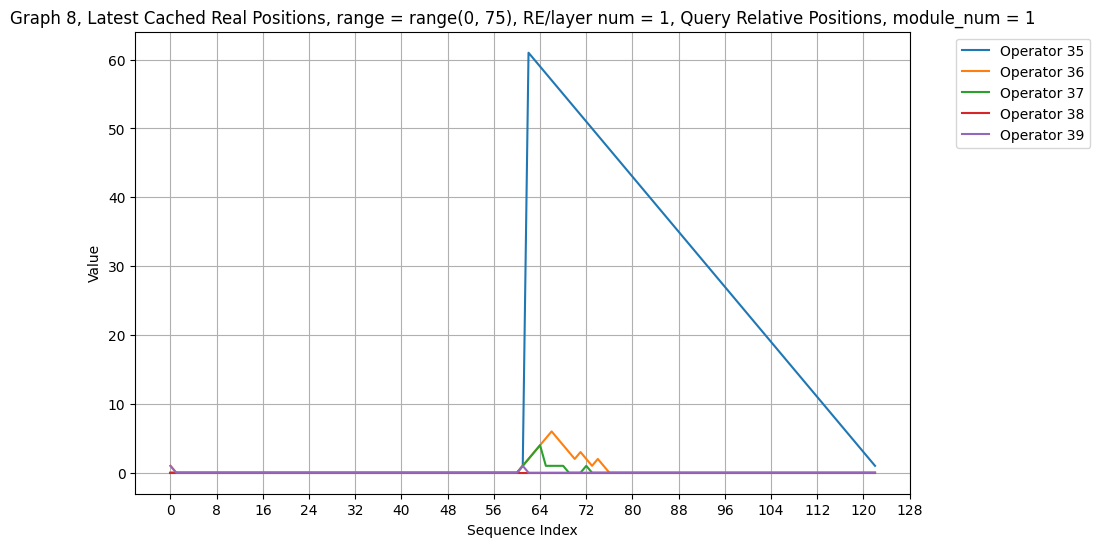}
  \caption{
  % \textbf{Figure A.5:}
  Reverse task. PRISM’s top-1 relative‐position mode via the blue cursor tracks the unique start symbol in the source string, enabling correct variable‐offset copying even at test lengths far from training.}
  \label{fig:reverse}
\end{figure}

\begin{figure}[H]
  \centering
  \includegraphics[width=\linewidth]{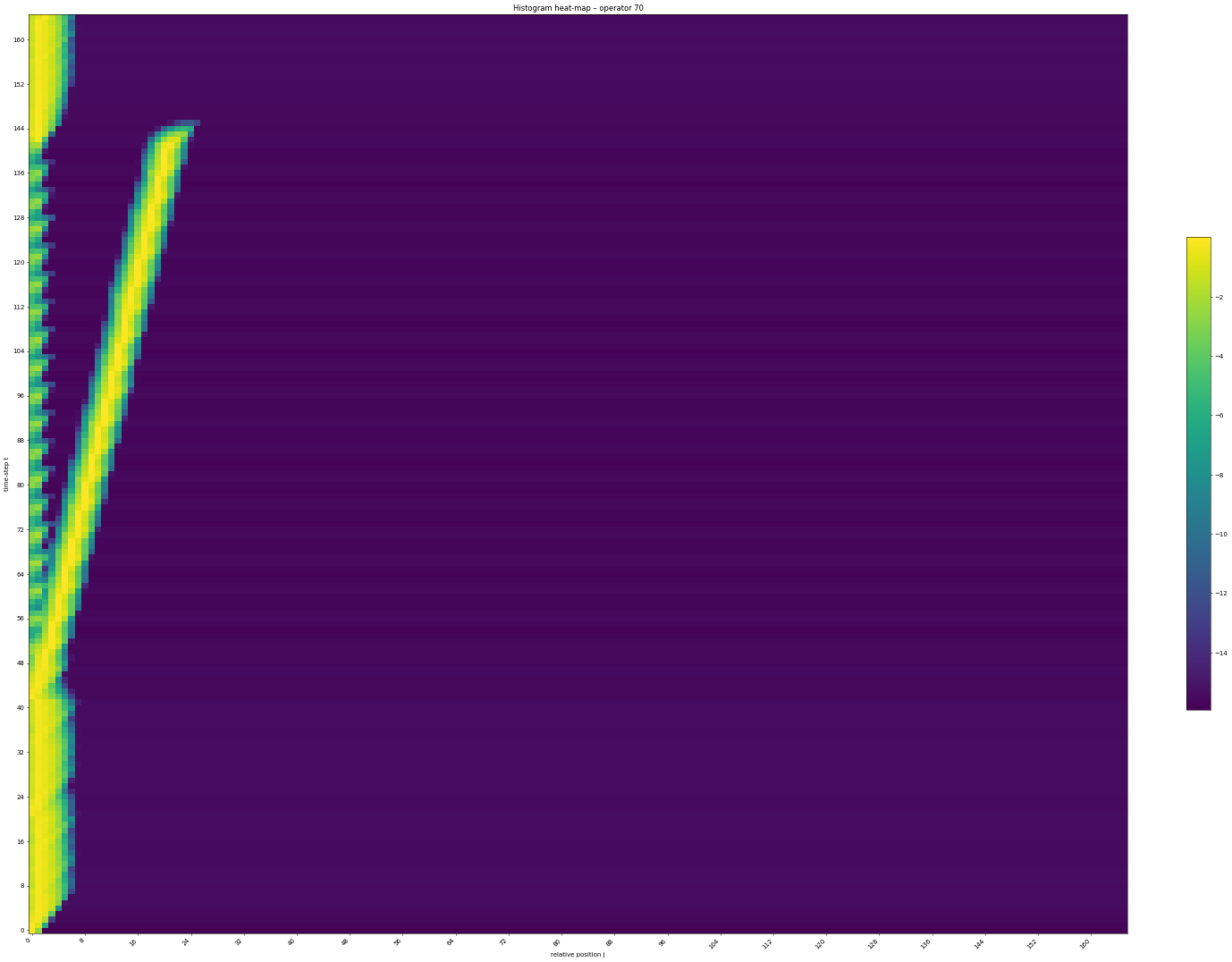}
  \caption{Addition (chain-of-thought)~(\ref{task:cot-addition}) task at OOD lengths. 
    The full histogram for each query token remains well-structured––peaked around the correct relative positions, attending to earlier digits \textit{via a non-1:1 mapping} to correctly write the chain-of-thought tokens––even when sequence lengths exceed those seen during training.}
  \label{fig:addition-ood}
\end{figure}

\begin{figure}[H]
  \centering
  \includegraphics[width=\linewidth]{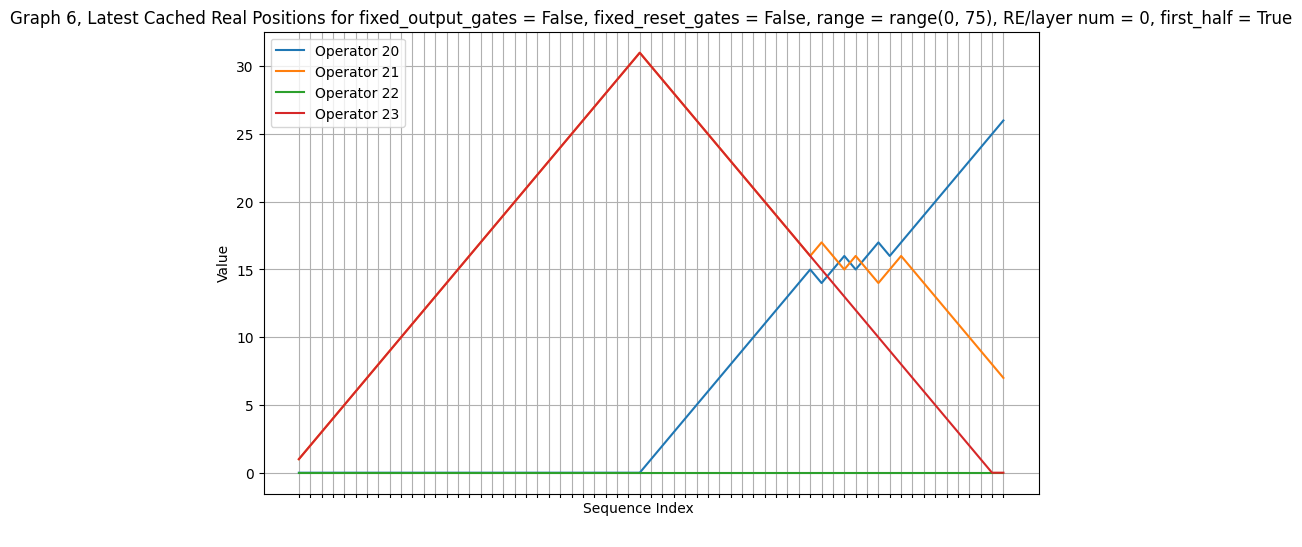}
  \caption{
  % \textbf{Figure A.7:} 
  Reverse‐string task.  The top-1 mode of the positional histogram reveals symmetric displacement patterns that exactly align with the positions required to reconstruct the reversed output, agnostic to input length. The query cursor histogram for operator (cursor) 23 demonstrates an \textit{emergent property} of the model, as both the query and key relative positions had to be learned in order to iterate through the input sequence in reverse.}
  \label{fig:reverse-queries}
\end{figure}

\begin{figure}[H]
  \centering
  \includegraphics[width=\linewidth]{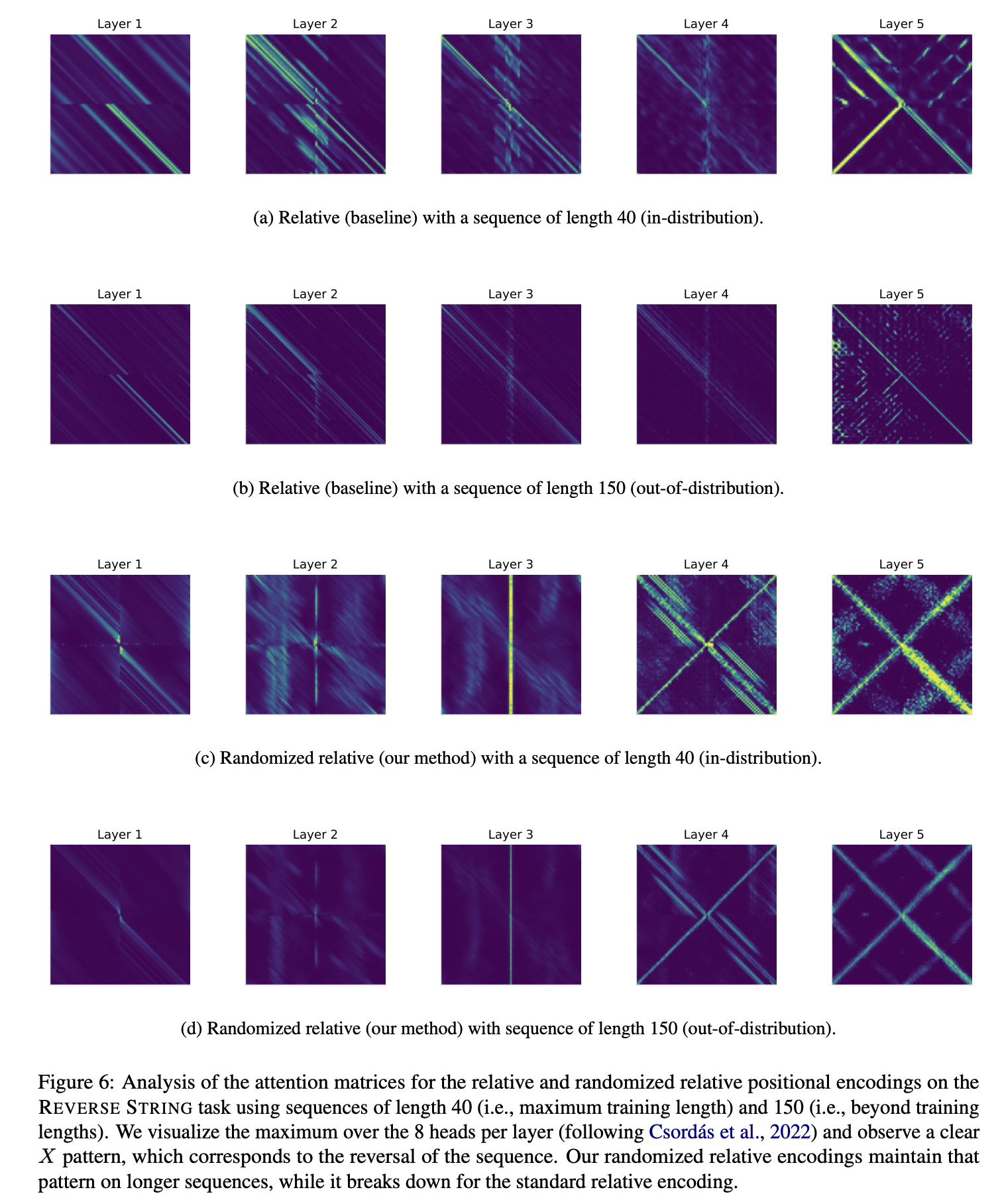}
  \caption{
  % \textbf{Figure A.8:} 
  Baseline comparison adapted from randomized PEs~\cite{Ruoss et al. 2023} on the reverse task.  Unlike PRISM’s learned histograms, randomized encodings fail to produce a coherent mode, demonstrating the advantage of data-driven positional distributions.}
  \label{fig:dm-randomized-pe}
\end{figure}

\begin{figure}[H]
  \centering
  \includegraphics[width=\linewidth]{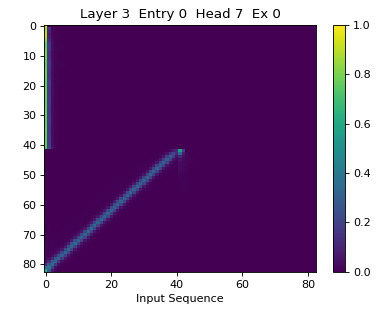}
  \caption{
  % \textbf{Figure A.9:} 
  Learned self-attention probabilities on the reverse task (in-distribution) for output sequence length 40 (the maximum trainable length, following Ruoss et al., 2023.  PRISM attends selectively to both local neighbors and distant tokens, capturing the two-sided dependencies needed for exact reversal.}
  \label{fig:reverse-mask}
\end{figure}

\begin{figure}[H]
  \centering
  \includegraphics[width=\linewidth]{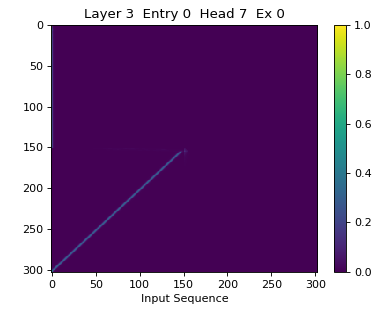}
  \caption{
  % \textbf{Figure A.10:} 
  Self-attention probabilities on the reverse task at an OOD length of 150. Unlike Ruoss et al., 2023, the same \textit{sparse}~\cite{golowich2025sparsity} attention pathways persist for much longer sequences, evidencing the model’s ability to generalize its positional reasoning far beyond the training horizon.}
  \label{fig:reverse-mask-ood}
\end{figure}

In this appendix we delve into the internal representations that PRISM
learns in order to achieve robust length extrapolation.  We begin by
showing the gate‐probability distributions at initialization and after
training (Figures~\ref{fig:init-gate-probs}-\ref{fig:trained-gate-probs}), which reveal how the model comes to favor
incremental cursor movements over resets or decrements.  Next, we
visualize the \emph{mode} of the learned positional histogram (the
top‐1 most likely offset) for each task, both at the start of training
and at convergence (Figures~\ref{fig:tokencopy-init}-\ref{fig:reverse}), to demonstrate how PRISM sharpens
its beliefs about relative positions.  We include out‐of‐distribution
(OOD) histograms for the addition and reverse tasks (Figures~\ref{fig:addition-ood}-\ref{fig:reverse-mask-ood})
to show that these learned distributions remain coherent far beyond the
training horizon. Finally, we display the self‐attention masks that
emerge on the reverse task (\ref{fig:reverse-mask}, \ref{fig:reverse-mask-ood}) and compare against a
randomized‐PE baseline (\ref{fig:dm-randomized-pe}), highlighting how our
differentially learned encodings outperform fixed or random schemes.

  % Now supply \footnotetext inside the same ‘figure’ so that it’s paired:
  \footnotetext{%
    “Keep” probabilities are implicitly represented here as
    $1 - (p_{\text{inc}} + p_{\text{dec}})$.%
  }

\section{Algorithmic Details}

\begin{algorithm}[H]
\caption{Histogram Update per Cursor \(c\) with Optional Copy Branch (Symmetric Offsets)}
\label{alg:histogram}
\begin{algorithmic}[1]
\Require For each cursor \(c\in\{1,\dots,C\}\): current histogram 
  \(\bm h_{t}^{(c)}\in\Delta^{L-1}\) \hfill(rows = indices \(0\ldots L-1\))
\Require Symmetric support: \(L = 2P+1\), center index \(\texttt{mid}=P\), where index \(i\) corresponds to offset \(d=i-\texttt{mid}\in\{-P,\dots,+P\}\)
\Require Gate scalars 
  \(p_{\mathrm{reset}}^{(c)},\,p_{\mathrm{incr}}^{(c)},\,p_{\mathrm{decr}}^{(c)},\,p_{\mathrm{keep}}^{(c)}\in[0,1]^{B}\)
\Require (Optional) copy distribution 
  \(\bm p_{\mathrm{copy}}^{(c)}\in\Delta^{L+1}\) 
  (last entry = “no-copy”)
\Ensure Next histogram \(\bm h_{t+1}^{(c)}\)

\Statex \textbf{(1) Aggregate mass}
\[
  \text{total}^{(c)} \;=\;\sum_{i=0}^{L-1} h_{t}^{(c)}[i]
  \quad\in\mathbb{R}^{B}
\]
\State Initialize \(\bm h_{t+1}^{(c)}\gets\bm 0\in\mathbb{R}^{L\times B}\)

\Statex \textbf{(2) Reset branch (re-center at offset 0, then take one step)}
\[
  \begin{aligned}
    h_{t+1}^{(c)}[\texttt{mid}+1] &\mathrel{+}= \text{total}^{(c)}\;\,p_{\mathrm{reset}}^{(c)}\;\,p_{\mathrm{incr}}^{(c)}
    \quad(\text{clamp at }L-1),\\
    h_{t+1}^{(c)}[\texttt{mid}]   &\mathrel{+}= \text{total}^{(c)}\;\,p_{\mathrm{reset}}^{(c)}\;\,p_{\mathrm{keep}}^{(c)},\\
    h_{t+1}^{(c)}[\texttt{mid}-1] &\mathrel{+}= \text{total}^{(c)}\;\,p_{\mathrm{reset}}^{(c)}\;\,p_{\mathrm{decr}}^{(c)}
    \quad(\text{clamp at }0).
  \end{aligned}
\]

\Statex \textbf{(3) No–reset branch (shift / keep on symmetric offsets)}
\[
  \begin{aligned}
    w_{\mathrm{inc}}^{(c)}   &= (1 - p_{\mathrm{reset}}^{(c)})\;p_{\mathrm{incr}}^{(c)},\\
    w_{\mathrm{dec}}^{(c)}   &= (1 - p_{\mathrm{reset}}^{(c)})\;p_{\mathrm{decr}}^{(c)},\\
    w_{\mathrm{keep}}^{(c)}  &= (1 - p_{\mathrm{reset}}^{(c)})\;p_{\mathrm{keep}}^{(c)},\\
    h_{t+1}^{(c)}[i]   &\mathrel{+}= h_{t}^{(c)}[i]\;\,w_{\mathrm{keep}}^{(c)},\\
    h_{t+1}^{(c)}[i+1] &\mathrel{+}= h_{t}^{(c)}[i]\;\,w_{\mathrm{inc}}^{(c)}\quad(\text{clamp at }L-1),\\
    h_{t+1}^{(c)}[i-1] &\mathrel{+}= h_{t}^{(c)}[i]\;\,w_{\mathrm{dec}}^{(c)}\quad(\text{clamp at }0).
  \end{aligned}
\]

\If{\(\bm p_{\mathrm{copy}}^{(c)}\) provided}
  \Statex \textbf{(4) Copy branch}
  \State Split \(\bm p_{\mathrm{copy}}^{(c)} = \bigl(\bm p_{\mathrm{pos}}^{(c)},\,p_{\mathrm{no\text{-}copy}}^{(c)}\bigr)\),
  where \(\bm p_{\mathrm{pos}}^{(c)}\in\mathbb{R}^{L\times B}\)
  \State Scale normal flows: 
    \(\bm h_{t+1}^{(c)}\;\times=\;p_{\mathrm{no\text{-}copy}}^{(c)}\)
  \State Add copied mass (per offset-bin \(k=0,\dots,L-1\)):
  \[
    h_{t+1}^{(c)}[k]\;\mathrel{+}=\;\bm p_{\mathrm{pos}}^{(c)}[k].
  \]
\EndIf

\Statex \textbf{(5) Sharpen \& renormalize}
\[
  \bm h_{t+1}^{(c)}
  \;\gets\;
  \frac{\bigl(\bm h_{t+1}^{(c)} + \varepsilon\bigr)^{\gamma_c}}
       {\sum_{i=0}^{L-1}\bigl(h_{t+1}^{(c)}[i]+\varepsilon\bigr)^{\gamma_c}},
  \quad
  \gamma_c>1
\]

\State \Return \(\bm h_{t+1}^{(c)}\)
\end{algorithmic}
\end{algorithm}

% \begin{algorithm}[H]
% \caption{\texttt{\_apply\_abs\_concat\_and\_ln\_optim}}
% \label{alg:abs‐ln‐optim}
% \begin{algorithmic}[1]
% \Require

\begin{algorithm}[H]
  %— Optional short caption for the “List of Algorithms” entry:
  \caption[Superposition Embedding per Cursor]
  {Efficient Superposition Embedding per Cursor:\\
   \texttt{\_get\_PE\_encoding\_vectors\_from\_scalar\_positions\_and\_norm}}
  \label{alg:superposition}
  \begin{algorithmic}[1]

\Require 
  Prob. tensors for queries \(\mathbf P^{(q)}\) and keys \(\mathbf P^{(k)}\)
  each of shape \(\R^{C\times B\times S\times r}\),
  where \(C = H\times O\) (number of cursors), \(r=\#\text{branches}\).\\
\Require Sinusoidal table \(\mathbf E\in\R^{r\times d}\).\\
\Ensure Embedded streams 
  \(\mathbf E_q,\mathbf E_k\in\R^{B\times H\times O\times S\times d}\).

% \For{\(X\) in \(\{\mathbf P^{(q)},\,\mathbf P^{(k)}\}\)}  
\For{\textbf{in parallel} $X\in\{\mathbf P^{(q)},\,\mathbf P^{(k)}\}$}

  \State {\footnotesize\(\triangleright\) Flatten out cursors, batch, sequence into one “row” dim}
  \[
    X_{\mathrm{flat}}
    \;\gets\;
    \mathrm{reshape}\bigl(X,\;[\,C\;B\;S,\;r\,]\bigr)
    \quad\in\R^{(C\,B\,S)\times r}
  \]
  \State {\footnotesize\(\triangleright\) One SIMD-style mat-mul over branches}\footnotemark
  \[
    Z_{\mathrm{flat}}
    \;\gets\;
    X_{\mathrm{flat}}\;\times\;\mathbf E
    \quad\in\R^{(C\,B\,S)\times d}
  \]
  \State {\footnotesize\(\triangleright\) Restore to \([C,B,S,d]\)}
  \[
    Z
    \;\gets\;
    \mathrm{reshape}\bigl(Z_{\mathrm{flat}},\;[\,C,\,B,\,S,\,d\,]\bigr)
  \]
  \State {\footnotesize\(\triangleright\) Turn \(C=H\times O\) back into head\,×\,cursor dims, then permute}
  \[
    Z
    \;\gets\;
    \mathrm{reshape}\bigl(Z,\;[\,H,\,O,\,B,\,S,\,d\,]\bigr)
    \;\xrightarrow{\;\mathrm{permute}(2,1,0,3,4)\;}
    \;\underbrace{\in\R^{\,B\times H\times O\times S\times d\,}}_{\mathbf E_{(\cdot)}}
  \]
  \State {\footnotesize\(\triangleright\) Exactly as in code’s \_apply\_abs\_concat\_and\_ln\_optim:}
  \[
    (\mathbf E_q,\mathbf E_k) 
      \;\gets\;
      \_apply\_abs\_concat\_and\_ln\_optim\bigl(
        Z,Z,\;B,H,O,S,d,\;\mathrm{abs\_enc},\;\mathrm{ablate\_abs\_pos\_enc}
      \bigr)
  \] 
  \If{$X = \mathbf P^{(q)}$}
    \State $\mathbf E_q \gets Z$
  \Else
    \State $\mathbf E_k \gets Z$
  \EndIf

\EndFor

\State \Return \(\mathbf E_q,\;\mathbf E_k\)
\end{algorithmic}
\end{algorithm}

\footnotetext{Instead of materializing $r$ separate $d_{\text{model}}$-dimensional
vectors for every branch and every token, we multiply once by the
$r\times d_{\text{model}}$ sinusoidal table (a single
\texttt{einsum}/GEMM).  The math is identical but the memory traffic is
$O(1/r)$ smaller, and the whole update stays in one SIMD-friendly
kernel.}

\subsection{Absolute-PE Concatenation \& Layer‐Norm Optimization}
\label{sec:abs‐ln‐optim}

This routine is called at the end of the superposition embedding to
(1) optionally concatenate an absolute positional embedding branch, and
(2) apply either a custom scaling or a projection‐free layer‐norm:

\begin{algorithm}[H]
\caption{\texttt{\_apply\_abs\_concat\_and\_ln\_optim}}
\label{alg:abs‐ln‐optim}
\begin{algorithmic}[1]
\Require
  Encodings \(\mathit{enc}_q,\mathit{enc}_k\in\R^{B\times H\times C\times S\times d}\),  \\
  absolute-PE table \(\mathit{abs\_enc}\in\R^{\ge S\times d}\),  
  flags \(\mathit{CONCAT\_ABS\_POSITION\_ENCODINGS},\linebreak\,\mathit{MHA\_USE\_CUSTOM\_LN}\),
  \(ablate\_flag \equiv False\)
\Ensure
  Updated \(\mathit{enc}_q,\mathit{enc}_k\) of same final shape.

\State \(C_{\text{base}}\gets C\)
\If{\(\mathit{CONCAT\_ABS\_POSITION\_ENCODINGS}\)}  
  \Comment{concatenate one extra absolute‐PE branch}
  \State \(C\gets C_{\text{base}}+1\)
  \State \(\mathit{abs\_pe}\gets \mathit{abs\_enc}[:S]\!\bigl[1,1,1,S,d\bigr]\!\to\!\mathrm{expand}(B,H,1,S,d)\)
  \If{ablate\_flag}
    \State \(\mathit{abs\_pe}\gets \mathbf{0}_{B\times H\times1\times S\times d}\)
  \EndIf
  \State \(\mathit{enc}_q\gets \mathrm{cat}(\mathit{enc}_q,\mathit{abs\_pe};\,\mathrm{dim}=2)\)
  \State \(\mathit{enc}_k\gets \mathrm{cat}(\mathit{enc}_k,\mathit{abs\_pe};\,\mathrm{dim}=2)\)
\EndIf

\If{\(\mathit{MHA\_USE\_CUSTOM\_LN}\)}  
  \Comment{global rescaling so norm = \(\sqrt d\)}
  \State \(\alpha\gets \texttt{sinusoidal\_encoding\_vector\_norm\_vs\_d\_model}(d)\)
  \State \(\mathit{scale}\gets \frac{\sqrt d}{\sqrt{C\cdot\alpha^2}}\)
  \State \(\mathit{enc}_q\;\boldsymbol\cdot=\;\mathit{scale},\quad  
         \mathit{enc}_k\;\boldsymbol\cdot=\;\mathit{scale}\)
\Else
  \Comment{mean-free LayerNorm across \([C,S]\) per head}
  \State assert \(\mathit{enc}_q,\mathit{enc}_k\text{ shaped }(B,H,C,S,d)\)
  \State \(\mathit{enc}_q\!\to\!\mathrm{permute}(0,1,3,2,4)\;[B,H,S,C,d]\)
  \State \(\mathit{enc}_k\!\to\!\mathrm{permute}(0,1,3,2,4)\;[B,H,S,C,d]\)
  \State 
    \(\mathit{enc}_q\gets
       \mathit{enc}_q/
       \bigl\|\mathit{enc}_q\bigr\|_{(-2,-1)}\times\sqrt d\)
  \State 
    \(\mathit{enc}_k\gets
       \mathit{enc}_k/
       \bigl\|\mathit{enc}_k\bigr\|_{(-2,-1)}\times\sqrt d\)
  \State 
    \(\mathit{enc}_q\!\to\!\mathrm{permute}(0,1,3,2,4),\quad
     \mathit{enc}_k\!\to\!\mathrm{permute}(0,1,3,2,4)\)
\EndIf

\State \Return \(\mathit{enc}_q,\;\mathit{enc}_k\)
\end{algorithmic}
\end{algorithm}

\begin{algorithm}[H]
\caption{Position-Aware Attention Scores 
(\_\!findPositionScores)}
\label{alg:compute-position-scores}

\begin{algorithmic}[1]
\Require
  Encodings 
  \(\mathit{enc}_q,\mathit{enc}_k\in\R^{B\times H\times C\times S\times d}\)
  \Comment{\(C\)=cursors per head}
\Require
  Learned scaling factors 
  \(\alpha\in\R^{H\times C}\) 
  (in code: \(\alpha\equiv\)\texttt{POST\_QK\_SCALING\_FACTORS\_MULTIPLIER}),\\
  \(\texttt{NEGATIVE\_COEFFS\_ALLOWED}=\texttt{false}\)
\Ensure
  \(\mathit{scores},\,\mathit{position\_scores}\in\R^{B\times H\times S\times S}\)

\State \((B,H,C,S,d)\gets \mathrm{shape}(\mathit{enc}_q)\)
\Comment{cursor index \(c=1,\dots,C\)}

\ForAll{\(b=1,\dots,B;\;h=1,\dots,H;\;c=1,\dots,C;\;i,j=1,\dots,S\)}
  \[
    \mathit{scores_{3}}[b,h,c,i,j]
    \;\gets\;
    \bigl\langle
      \mathit{enc}_q[b,h,c,i,:],\,
      \mathit{enc}_k[b,h,c,j,:]
    \bigr\rangle
  \]
\EndFor

\Statex\Comment{\textbf{Post-scaling (always nonnegative):}}
\State \(\displaystyle
  \beta[h,c]\;\gets\;|\alpha[h,c]|
\)
\State \(\displaystyle
  \mathit{scores_{3}}[b,h,c,i,j]\;\;\boldsymbol\cdot=\;\beta[h,c]
\)

\Statex\Comment{\textbf{Sum out cursors:}}
\[
  \mathit{scores_{2}}[b,h,i,j]
    \;=\;\sum_{c=1}^{C}\mathit{scores_{3}}[b,h,c,i,j]
\]

\State \(N\;\gets\;C \times d\)

\Statex\Comment{\textbf{Final normalization:}}
\[
  \mathit{scores_{final}}[b,h,i,j]
    \;=\;
    \frac{\mathit{scores_{2}}[b,h,i,j]}{\sqrt{N}}
\]
\State \(\mathit{position\_scores}\gets \mathit{scores_{final}}\)

\State \Return \(\mathit{scores},\,\mathit{position\_scores}\)
\end{algorithmic}
\end{algorithm}

\section{SCAN-CoT Length Distributions}\label{sec:scan-len-dataset-distr}

Here we show histograms of token‐length distributions for the SCAN dataset.

\begin{figure}[H]
  \centering
  \includegraphics[width=0.8\textwidth]{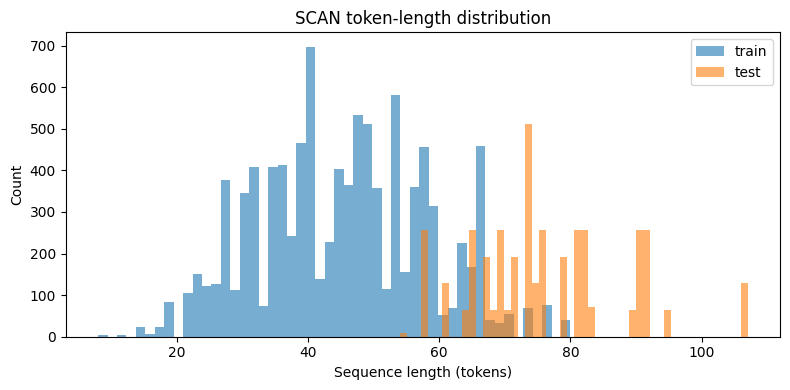}
  \caption{Full‐sequence lengths, SCAN train and test split.}
  \label{fig:scan-full-seq-lens}
\end{figure}

\begin{figure}[H]
  \centering
  \includegraphics[width=0.8\textwidth]{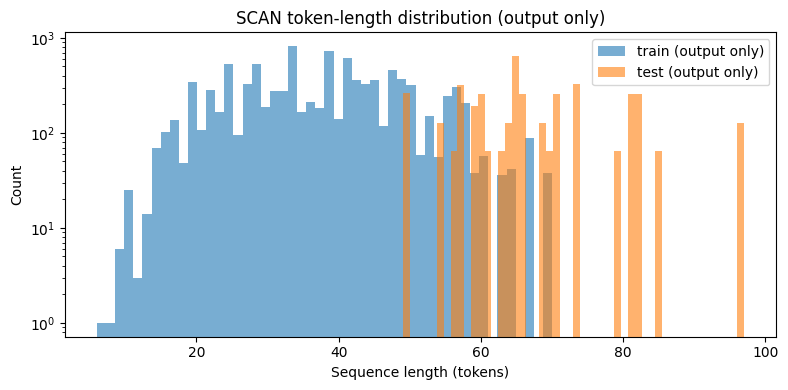}
  \caption{Output‐only lengths (tokens right of ``='') on a log scale.}
  \label{fig:scan-output-only}
\end{figure}

\begin{figure}[H]
  \centering
  \includegraphics[width=0.8\textwidth]{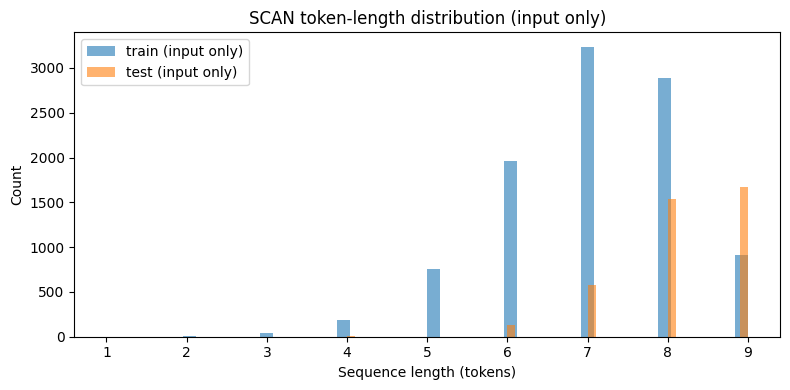}
  \caption{Input‐only lengths (tokens left of ``='').}
  \label{fig:scan-input-only}
\end{figure}

\newpage
% \section*{References}

% \begin{thebibliography}{99}

% \section*{References}
% \begin{thebibliography}{99}
\nocite{*}

\printbibliography[title={References}]

\end{document}